\documentclass[10pt,twocolumn,letterpaper]{article}
\usepackage[arxiv]{iccv}
\pdfoutput=1
\definecolor{iccvblue}{rgb}{0.21,0.49,0.74}
\usepackage[pagebackref,breaklinks,colorlinks,allcolors=iccvblue]{hyperref}
\def\paperID{2907} 
\def\confName{ICCV}
\def\confYear{2025}
\definecolor{cvprblue}{rgb}{0.21,0.49,0.74}
\usepackage{times}
\usepackage{latexsym}
\usepackage[T1]{fontenc}
\usepackage[utf8]{inputenc}
\usepackage{microtype}
\usepackage{amsmath}
\usepackage{inconsolata}
\usepackage{graphicx}
\usepackage{caption}
\usepackage{multirow}
\usepackage[normalem]{ulem}
\useunder{\uline}{\ul}{}
\usepackage{tikz}
\usepackage{float}
\usepackage{adjustbox}
\usepackage{subcaption} 
\usepackage{enumitem}
\usepackage{booktabs}

\definecolor{myHexPurple}{HTML}{9370DB} 
\title{
Exploring Hallucination of Large Multimodal Models in Video Understanding: Benchmark, Analysis and Mitigation}
\renewcommand\footnotemark{}
\author{
\!Hongcheng Gao$^{*1}$, Jiashu Qu$^{*2}$, Jingyi Tang$^{*1,3}$, Baolong Bi$^{1}$, Yue Liu$^{4}$, Hongyu Chen$^{5}$ \\ Li Liang$^{\dagger1,3}$, Li Su$^{\dagger1}$, Qingming Huang$^{1,3}$   \thanks{$^*$Equal contribution.}
\thanks{$^{\dagger}$Correspondence to Li Su and Liang Li.}\\
  $^{1}$University of Chinese Academy of Sciences $^{2}$University of Cincinnati \\
  $^{3}$Key Lab of Intell. Info. Process., Inst. of Comput. Tech., CAS \\$^{4}$National University of Singapore $^{5}$Beijing Jiaotong University \\
\normalsize{\texttt{\{gaohongcheng23,tjy23\}@mails.ucas.ac.cn}; \texttt{quju@mail.uc.edu}}  
}

\begin{document}
\maketitle
\begin{abstract}
\vspace{-0.68cm}

The hallucination of large multimodal models (LMMs), providing responses that appear correct but are actually incorrect, limits their reliability and applicability. This paper aims to study the hallucination problem of LMMs in video modality, which is dynamic and more challenging compared to static modalities like image and text. From this motivation, we first present a comprehensive benchmark termed \textcolor{myHexPurple}{\textbf{HAVEN}} for evaluating hallucinations of LMMs in video understanding tasks. It is built upon three dimensions, i.e., hallucination causes, hallucination aspects, and question formats, resulting in 6K questions. Then, we quantitatively study 7 influential factors on hallucinations, e.g., duration time of videos, model sizes, and model reasoning, via experiments of 16 LMMs on the presented benchmark. In addition, inspired by recent thinking models like OpenAI o1, we propose a video-thinking model to mitigate the hallucinations of LMMs via supervised reasoning fine-tuning (SRFT) and direct preference optimization (TDPO)—where SRFT enhances reasoning capabilities while TDPO reduces hallucinations in the thinking process. Extensive experiments and analyses demonstrate the effectiveness. Remarkably, it improves the baseline by 7.65\% in accuracy on hallucination evaluation and reduces the bias score by 4.5\%.
The code and data are public at \url{https://github.com/Hongcheng-Gao/HAVEN}.

\end{abstract}    
\vspace{-0.2cm}
\section{Introduction}
\label{sec:intro}

\begin{figure*}[t!]
\begin{center}
\vspace{-0.2cm}
\includegraphics[width=0.92\linewidth]{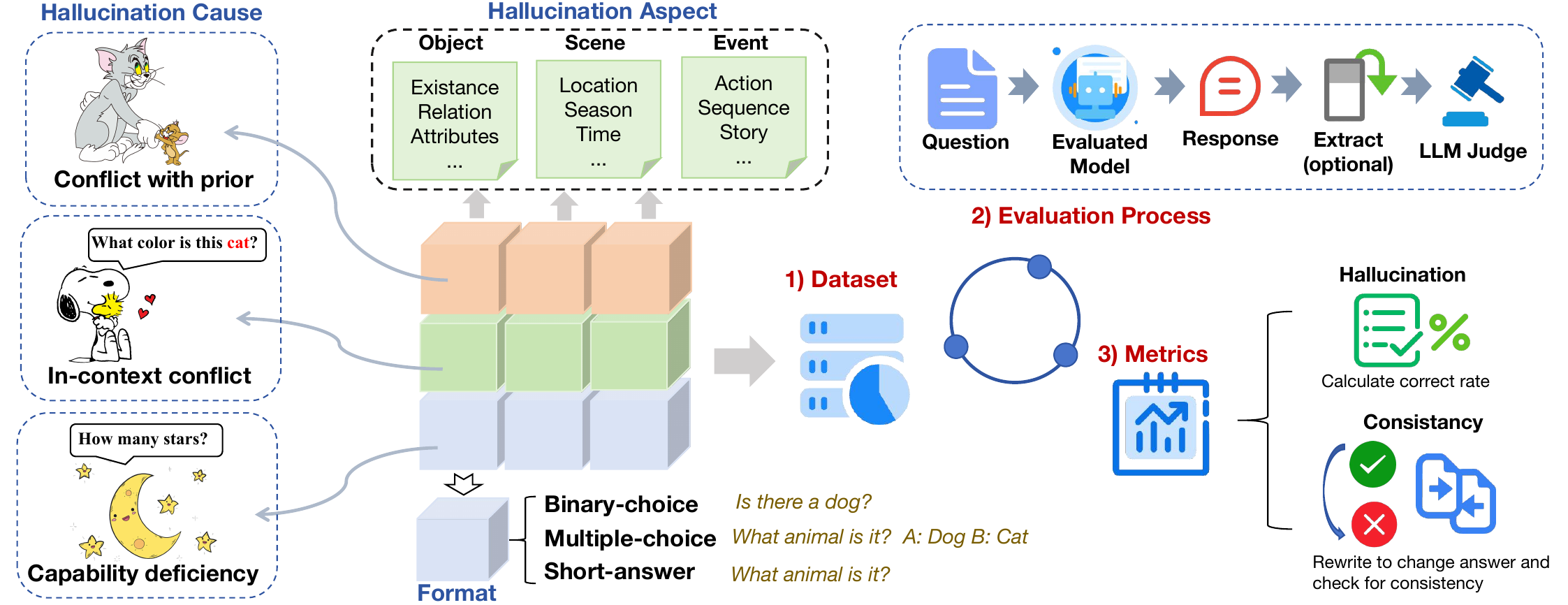}
\end{center}
\vspace{-0.4cm}
   \caption{\textbf{Construction protocol of HAVEN.} The left section outlines the three dimensions of data construction and the associated categories within each, while the right section details the evaluation process and metrics.} 
    \label{protocal}
    \vspace{-0.4cm}
\end{figure*}


In recent years, the success of Large Language Models (LLMs)~\cite{touvron2023llama,touvron2023llama2,jiang2024mixtral,jiang2023mistral} has significantly advanced the development of Large Multimodal Models (LMMs)\cite{yin2023survey,LiuLWL23a,openai2023gpt,claude3,reid2024gemini,liu2023improved,NEURIPS2023_9a6a435e}. By integrating vision and audio modules with LLMs, LMMs are capable of generating accurate responses based on users' textual prompts, visual inputs, and even audio data~\cite{shao2023prompting,sammani2022nlx,fu2023mme,zhou2024navgpt,gupta2023visual}. However, LLMs are found to suffer from hallucinations—\emph{providing responses that appear correct but are actually incorrect}—especially when faced with questions that conflict with their training data~\cite{li2023halueval, zhang2023siren}. Despite the advancements in LMMs, they also experience similar hallucination issues in both text and image understanding~\cite{huang2023survey,ji2023survey,DaiLJSF23}. This phenomenon undermines user trust in LMMs and limits their applicability in high-stakes areas such as healthcare~\cite{liu2023medical} and autonomous driving~\cite{tian2024drivevlm}. In response to these challenges, a series of studies have been proposed to evaluate and mitigate hallucinations in both LLMs and LMMs.



However, previous research on hallucinations of LMMs has primarily focused on image understanding~\cite{LiDZWZW23,liu2024phd,liu2023hallusionbench,jiang2024hal,lovenia2023negative,huang2024visual,kaul2024throne,fieback2024metatoken}, as earlier LMMs could not process video inputs. These benchmarks are designed to evaluate hallucinations involving factors such as objects, relationships and attributes in a single image. With advancements in multimodal technologies, numerous LMMs now support video processing. Although many of these models did not incorporate audio inputs from videos, most can effectively process the visual content of video. Unlike image understanding, videos consist of sequences of multiple image frames over time, making video understanding more complex. It requires the analysis of continuous temporal dynamics, including sequential changes in human actions, object movements, and scene transitions. Hence, hallucinations in video understanding also differ from those in images.

To address the concern above, we first proposed a benchmark for \textbf{HA}llucination in \textbf{V}ideo Und\textbf{E}rsta\textbf{N}ding (\textcolor{myHexPurple}{\textbf{HAVEN}}). HAVEN is meticulously designed to quantitatively evaluate the hallucination in video understanding for LMMs, which is constructed based on the following dimensions: (i) Three \emph{causes} of hallucinations: conflict with prior knowledge, in-context conflict, and inherent capability deficiencies of LMMs. (ii) Three types of hallucination \emph{aspects} in a video: object, scene, and event. (iii) Three \emph{formats} of questions: binary-choice, multiple-choice, and short-answer.

Our evaluation of 16 existing LMMs reveals that Valley-Eagle-7B and GPT4o-mini demonstrated the lowest hallucination rates among all models tested, while Qwen2.5-VL-3B and Valley-Eagle-7B exhibited superior response consistency. The analysis further indicated that model performance is influenced by several factors: accuracy initially increases then decreases with longer video duration, decreases with more complex questions, improves with increased frame sampling, and generally shows reduced hallucinations and higher consistency with larger model sizes. Besides, chain-of-thought reasoning can reduce hallucinations in LMMs.

In addition, we propose a thinking-based training strategy to reduce hallucinations by enhancing the LMM's reasoning abilities. This training strategy is divided into two steps: supervised reasoning fine-tuning (SRFT) and thinking-based direct preference optimization (TDPO). In the SRFT stage, we perform supervised fine-tuning on the LMM using videos derived from images, incorporating long Chain-of-Thought answers distilled from image-thinking models like QVQ~\cite{qvq-72b-preview} and OpenAI o1 to equip the model with thinking capabilities. The TDPO stage directly optimizes the fine-grained thinking component at both the word and sentence levels, with fabricated reasoning receiving stronger feedback to ensure it remains factually grounded. Experimental results indicate that our training methodology yields substantial improvements in both reducing hallucinations and enhancing response consistency in LMM, as exemplified by LLaVA-NEXT-Video-DPO-7B.


\vspace{-0.1cm}
\section{Related Works}
\label{sec:rel_works}
\begin{figure*}[t!]
\centering
    \centering
    \begin{subfigure}[b]{0.328\textwidth}
        \centering
        \includegraphics[width=\textwidth]{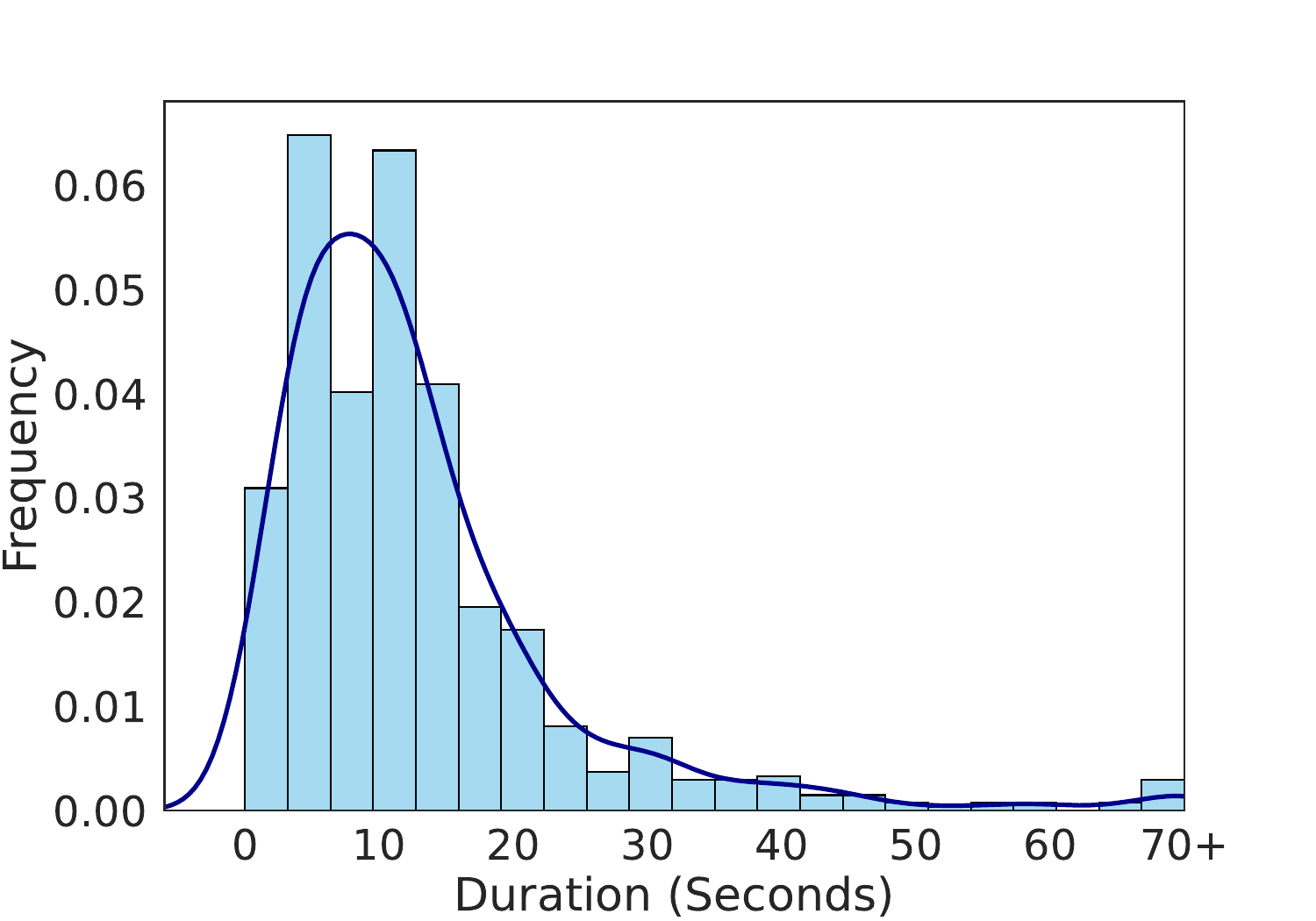}
        \caption{Duration Time}
        \label{hit:subfig1}
        \vspace{-0.2cm}
    \end{subfigure}
    \hfill
    \begin{subfigure}[b]{0.328\textwidth}
        \centering
        \includegraphics[width=\textwidth]{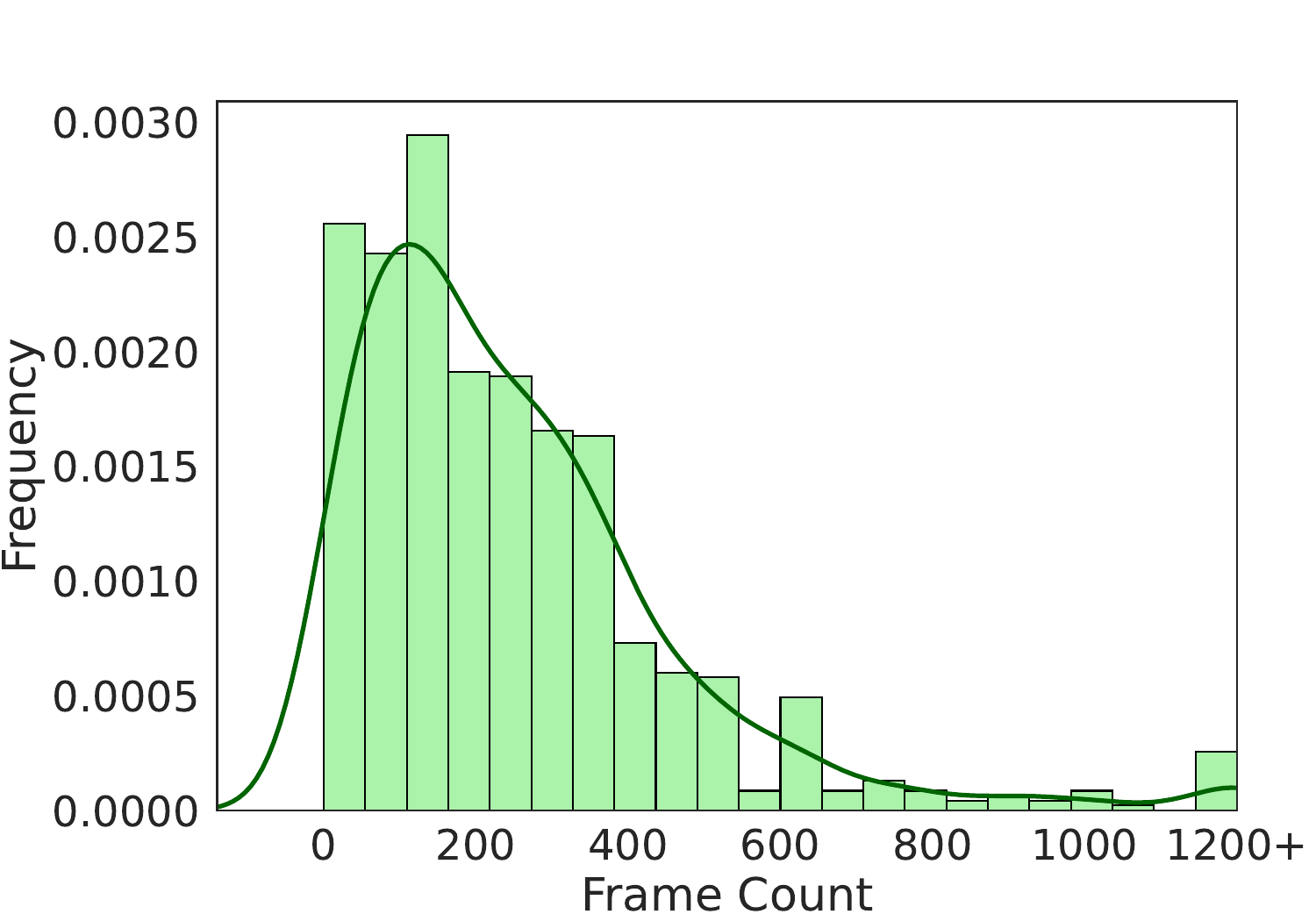}
        \caption{Frame Count}
        \label{hit:subfig2}
        \vspace{-0.2cm}
    \end{subfigure}
    \hfill
    \begin{subfigure}[b]{0.328\textwidth}
        \centering
        \includegraphics[width=\textwidth]{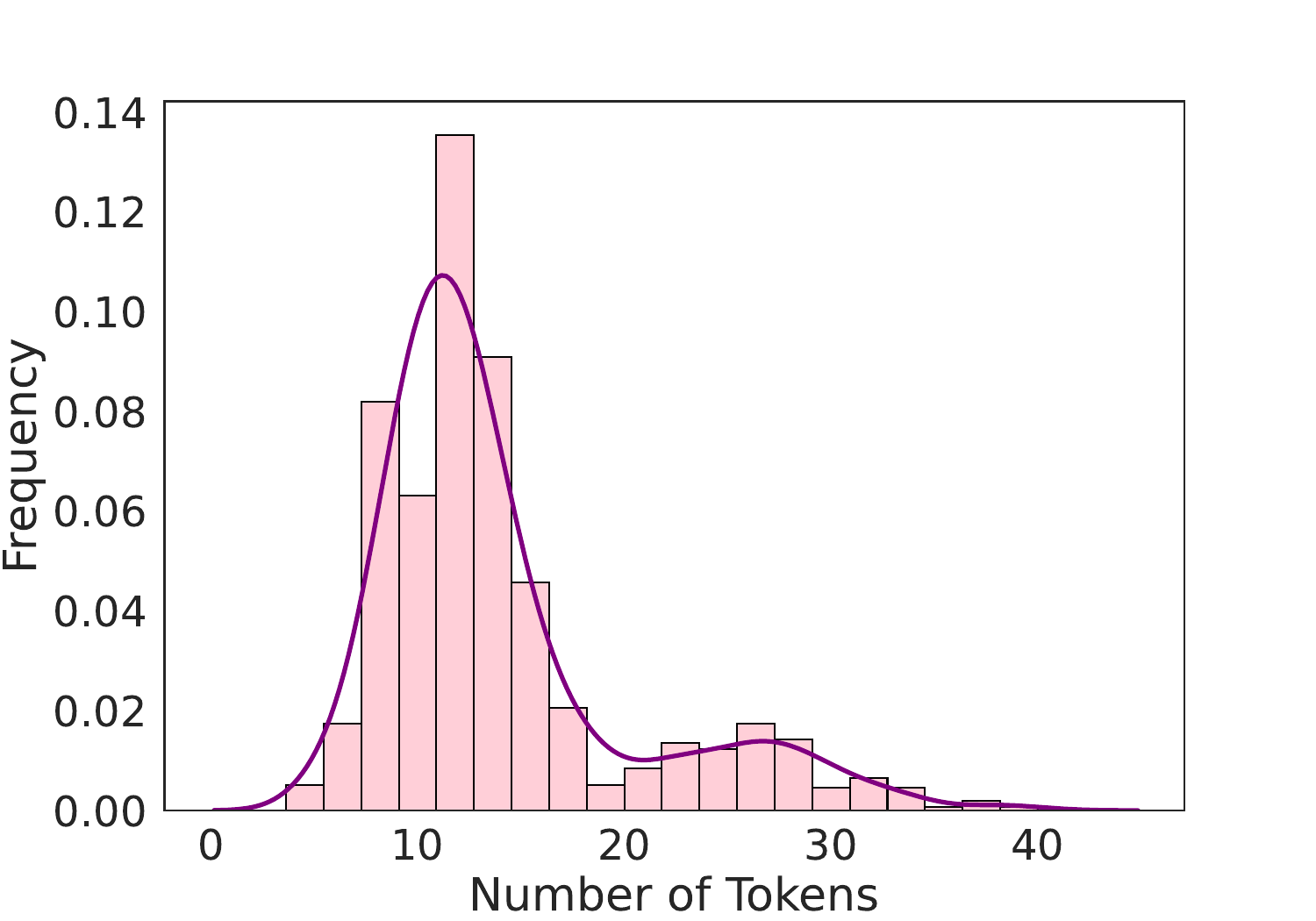}
        \caption{Question Length}
        \label{hit:subfig3}
        \vspace{-0.2cm}
    \end{subfigure}
    \caption{Distribution of duration time, frame count, and question length.}
    \label{fig:distri}
    \vspace{-0.3cm}
\end{figure*}

\begin{figure}[t!]
\begin{center}
\includegraphics[width=0.8\linewidth]{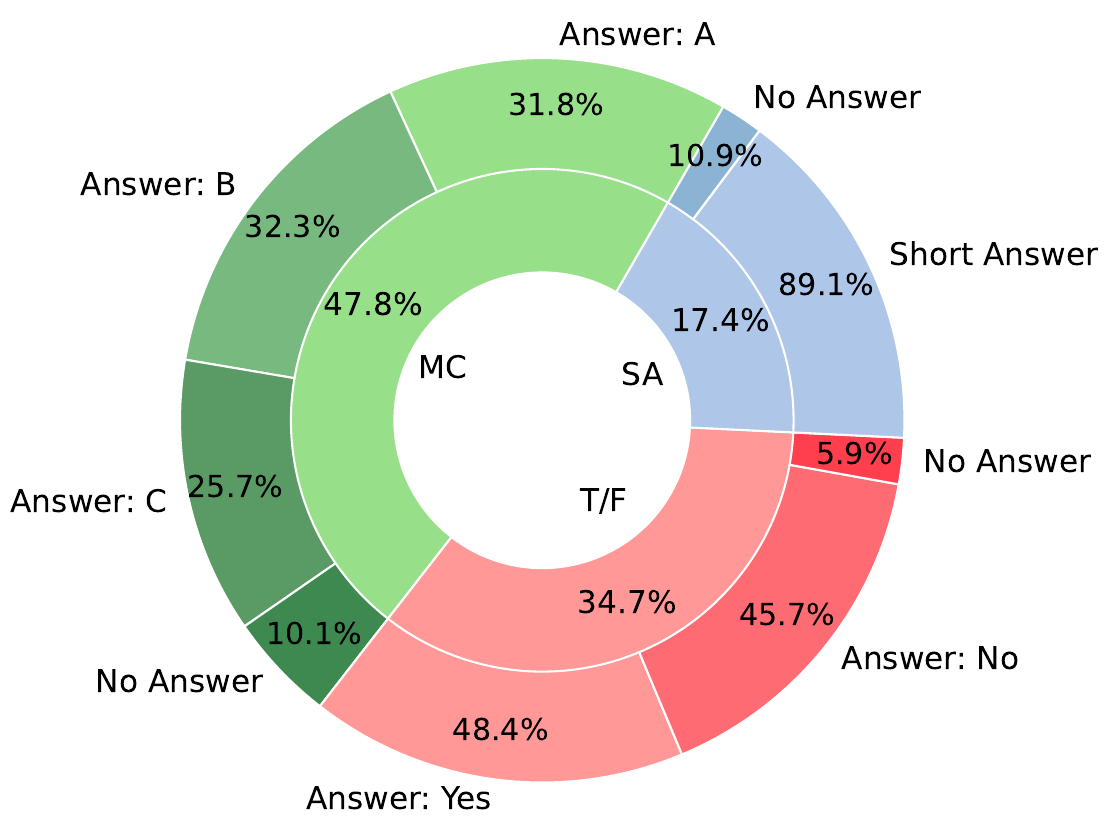}

\end{center}

\vspace{-0.5cm}
   \caption{\textbf{Question format distribution.} Percentage share of each format-binary-choice (T/F), multiple-choice (MC), and short-answer (SA)—and the proportion occupied by the detailed answer.} 
   \vspace{-0.5cm}
    \label{data_format}
\end{figure}
\vspace{-0.1cm}
\paragraph{Large Multimodal Models (LMMs).}
Recent LMMs primarily focused on visual-language understanding, which can be categorized into image-language~\cite{LiuLWL23a,openai2023gpt,claude3,liu2023improved} and video-language models~\cite{lin2023video,maaz2023video,zhang2023video,reid2024gemini}. Early visual-language models, designed for single-image inputs~\cite{radford2021learning,jia2021scaling,li2019visualbert,kim2021vilt}, followed two main approaches: self-supervised learning with image-text pairs~\cite{radford2021learning,jia2021scaling}, and adding image adapters to existing LLMs followed by alignment fine-tuning~\cite{li2019visualbert,kim2021vilt}. 
Subsequent research extended to multi-image processing capabilities, naturally evolving into video-language models, which leverage frame-by-frame processing inherited from image-language models to achieve video understanding through text-video alignment~\cite{lin2023video,maaz2023video,zhang2023video,zhang2024llavanext-video,xu2024pllava,li2024llamavid,cheng2024videollama,chen2024sharegpt4video}. VideoLLaVA~\cite{lin2023video}, for example, incorporates a LanguageBind encoder to align visual features from both images and videos into a unified space for joint video-text encoding. Recent developments have further expanded to modalities including audio~\cite{cheng2024videollama,huang2025step} and 3D point clouds~\cite{xu2025pointllm}.

\vspace{-0.4cm}
\paragraph{Hallucination Benchmarks.} Hallucination is widespread in large models~\cite{touvron2023llama,touvron2023llama2,jiang2024mixtral,jiang2023mistral}, and the relevant benchmarks primarily target LLMs and VLMs. Hallucination benchmarks for LLMs\cite{han2019episodic,li2023halueval,lin2021truthfulqa,mckenzie2023inverse,wei2024measuring,ravichander2025halogen} mainly focus on identifying factual inaccuracies, deviations from original contexts and the self-awareness. 
For VLM, hallucination refers to the phenomenon that the generated text responses are inconsistent with the corresponding visual content~\citep{bai2024hallucination}. Numerous benchmarks~\cite{LiDZWZW23,liu2024phd,liu2023hallusionbench,jiang2024hal,lovenia2023negative,huang2024visual,kaul2024throne,fieback2024metatoken,jing2023faithscore,cui2023holistic,wang2023llm,wang2024mitigating,chen2024unified,cha2024visually,sun2023aligning} have been developed to evaluate these hallucinations. 
Early benchmarks (e.g., CHAIR~\cite{rohrbach2018object} and POPE~\cite{li2023evaluating}) primarily target object presence, while later ones like HallusionBench~\cite{guan2023hallusionbench} and MMHal-Bench~\cite{sun2023aligning} also evaluate object relationships and scene understanding. Recent studies~\cite{jing2023faithscore,wang2024hawkeye,liu2024phd,cui2023holistic} have begun addressing factual hallucinations in image-language models, yet most approaches focus on single images, neglecting continuous content. 
VideoHallucer~\cite{wang2024videohallucer} bridges this gap by introducing a video hallucination benchmark, though current methods still tend to consider only one hallucination dimension or treat multiple dimensions at the same level, leading to category overlaps.
\paragraph{Hallucination Mitigation.} 
Approaches to mitigating hallucinations can be broadly classified into two categories based on the stage at which they intervene. The first category deals with the training phase, which focuses on creating datasets specifically for
hallucination-related tasks~\cite{gunasekar2023textbooksneed,gunjal2024detectingpreventinghallucinationslarge,liu2024mitigatinghallucinationlargemultimodal,yu2024hallucidoctormitigatinghallucinatorytoxicity,tang2025mitigating,Shen_2021,raunak2021curiouscasehallucinationsneural} and developing novel training objectives or methodologies aimed at mitigating hallucinations~\cite{chen2024alleviatinghallucinationslargevisionlanguage,lee2023factualityenhancedlanguagemodels,shi2024incontextpretraininglanguagemodeling}. 
Although these methods have demonstrated effectiveness, they often require extensive training processes, making them both time-consuming and resource intensive. The second category focuses on the inference phase, usually involves CD-based~\cite{li2023contrastivedecodingopenendedtext} novel decoding
strategies~\cite{chen2024halcobjecthallucinationreduction,chuang2024doladecodingcontrastinglayers,kim2024codecontrastingselfgenerateddescription,liu2024payingattentionimagetrainingfree,phan2024distillationcontrastivedecodingimproving,kim2024vacodevisualaugmentedcontrastive,woo2024dontmissforesttrees,park2024conviscontrastivedecodinghallucination,zhao2024enhancingcontextualunderstandinglarge,zhu2024multilingualcontrastivedecodinglanguageagnostic}. Another approach is to detect potential hallucinations while
generating and correct them~\cite{kuhn2023semanticuncertaintylinguisticinvariances,nikitin2024kernellanguageentropyfinegrained}, among others~\cite{li2024inferencetimeinterventionelicitingtruthful,yue2024moremitigatingmultimodalhallucination,woo2024ritualrandomimagetransformations,zhang2024truthxalleviatinghallucinationsediting}. In addition, some researchers tackle hallucinations by manipulating attention weight assigned
to the image~\cite{zhu2024ibdalleviatinghallucinationslarge,zhang2023onenetenhancingtimeseries,xiao2024seeingimageprioritizingvisual,huo2024selfintrospectivedecodingalleviatinghallucinations,wu2024noiseboostalleviatinghallucinationnoise,li2025treblecounterfactualvlmscausal}, and prompt-based
methods~\cite{han2024skipnsimplemethod,xu2024rereadingimprovesreasoninglarge,wang2024mitigatinghallucinationslargevisionlanguage,qu2024lookcomparedecidealleviating,wu2024logicalclosedloopuncovering,wang2024videohallucer}. 
\section{\textcolor{myHexPurple}{HAVEN} Benchmark}
\label{sec:bench}

\subsection{Data Sources}
Our video data comprises videos from three public video datasets (COIN~\cite{tang2019coin}, ActivityNet~\cite{caba2015activitynet}, and Sports1M~\cite{kang2016review}) and 260 manually collected video clips from YouTube\footnote{https://www.youtube.com/}.

\subsection{Construction Protocol}
As shown in Fig.~\ref{protocal}, we first constructed a dataset based on three dimensions—\emph{hallucination causes, hallucination aspects, and question formats}—before implementing an automated evaluation process along with two metrics.

\subsubsection{Hallucination Causes}
We categorize hallucinations in LMMs into three types based on their underlying causes: conflicts with prior knowledge, in-context conflicts, and inherent capability deficiencies.

\vspace{-0.4cm}
\paragraph{Conflict with Prior.} 
When the content extracted from the video contradicts the model's inherent prior knowledge acquired during training, the model may rely on its prior knowledge learned from established facts or widely accepted knowledge, leading to incorrect responses that don't align with the provided information.
\vspace{-0.4cm}
\paragraph{In-context Conflict.} 
When task components conflict—such as discrepancies between the question and options or between the video and the questions/answers—valid answers cannot be derived from the given material. In such instances, the model should indicate uncertainty by responding with "I don’t know/no answers". However, models prone to hallucination often produce fabricated responses by focusing on limited contextual cues.

\vspace{-0.2cm}
\paragraph{Capability Deficiencies.} 
Most LMMs struggle with numerical tasks, often leading to miscounts or quantification errors in videos. For example, they might label a scene as having four cars when there are actually three, or miscount the number of people present. Such inaccuracies can undermine the reliability of video analysis by providing incorrect quantitative details.




\subsubsection{Hallucination Aspects}
According to the components of a video, hallucination aspects can be categorized into three types: object, scene, and event hallucinations.
\paragraph{Object.} Object hallucinations involve inaccuracies or fabrications related to the entities present in the video. This primarily includes the presence of objects, their inherent attributes such as size, color, and shape, as well as the emotions displayed by characters.
\vspace{-0.4cm}
\paragraph{Scene.} Scene hallucinations involve inaccuracies related to the overall setting and environment of the video. This includes errors concerning the background, location, season, time of day, lighting conditions, and the general context in which events take place. For example, a scene-level hallucination might mistakenly depict an indoor setting as outdoor, alter the time of day, or misrepresent environmental factors such as weather conditions.
\vspace{-0.4cm}
\paragraph{Event.} Event hallucinations involve distortions or false representations of the actions and occurrences within the video. This includes inaccuracies in the sequence of events, the nature of interactions between characters, and the specific actions taking place. For instance, an event-level hallucination might depict a character performing an action that does not actually occur in the video or misrepresent the timing and order of events.


\subsubsection{Question Types}
We classify questions into three types: binary-choice(T/F), multiple-choice(MC), and short-answer(SA). Binary-choice questions are phrased as interrogative sentences (e.g., "Is there a dog?") and require a response of "yes" or "no." Multiple-choice questions present a set of candidate answers, while short-answer questions call for a concise response. If a question lacks a definitive answer, the model should respond with "no answer" or "I don't know." 

\subsection{Data Post-processing}
To evaluate the inherent response biases of language models—namely their predisposition toward particular responses in binary-choice questions and multiple-choice questions—we implement a systematic question transformation protocol. For each binary question, we generate two complementary variants: one where "yes" is the correct answer and another where "no" is correct. Similarly, we transform each multiple-choice question into three parallel versions, with each version designating a different option (A, B, or C) as the correct answer. This approach aims to neutralize potential positional biases in model responses.

\subsection{Dataset Statistics}
\begin{table}[ht]

\centering
\resizebox{0.48\textwidth}{!}{%
\begin{tabular}{c|cccc}
\toprule
\textbf{Cause/Aspect} & \textbf{Object} & \textbf{Scene} & \textbf{Event} & \textbf{\#Total} \\ 
\midrule
\textbf{Prior Conflict}       & 2162 & 686  & 1763 & 4569 \\
\textbf{In-context Conflict} & 94   & 82   & 404  & 538  \\
\textbf{Capability}          & 1107 & 121  & 78   & 1156 \\
\textbf{\#Total}             & 3363 & 889  & 2245 & 6497 \\ 
\bottomrule
\end{tabular}%
}
\vspace{-0.2cm}
\caption{The Statistics of HAVEN}
\label{statistic_all}
\vspace{-0.4cm}
\end{table}




As shown in Table~\ref{statistic_all} and Fig.~\ref{data_format}, our dataset comprises 6,497 questions. The three question formats are distributed approximately in a 5:2:3 ratio. Among the three types of hallucinations, the "prior conflict" category accounts for the largest share—around 75\%—while the three aspects are proportioned roughly as 3:1:2.

Figure~\ref{fig:distri} shows the distribution of our data in terms of duration, frame count, and question length. The duration spans from 0 to 70 seconds, with most instances falling within the 0–20 second interval. The frame count ranges from 0 to 1200 frames, with the majority concentrated between 0 and 400 frames. The question length is computed based on the token count obtained via GPT4 tokenizer, with a dominant range of 10–15 tokens.

\begin{table*}[t]
\vspace{-0.4cm}
\centering
\label{hallucination}
\resizebox{0.98\textwidth}{!}{
\begin{tabular}{c|cccccccccc}
\toprule
\multirow{2}{*}{\textbf{Model/Type}} & \multicolumn{3}{c}{\textbf{Prior}} & \multicolumn{3}{c}{\textbf{In-context}} & \multicolumn{3}{c}{\textbf{Capability}} & \multirow{2}{*}{\textbf{Total}} \\
\cmidrule(lr){2-4} \cmidrule(lr){5-7} \cmidrule(lr){8-10}
                                     & Object         & Scene          & Event          & Object         & Scene          & Event          & Object         & Scene          & Event          &                                   \\ 
\midrule
VideoChatGPT-7B\citep{Maaz2023VideoChatGPT}                         & 34.78          & 43.88          & 38.00          & 17.02          & 13.41          & 17.57          & 32.52          & 47.11          & 20.51          & 34.69                             \\
Valley-Eagle-7B \citep{wu2025valley2}                      & \textbf{68.55} & \textbf{75.95} & \textbf{63.52} & 24.47          & 43.90          & 15.10          & {\ul 57.36}    & \textbf{55.37} & {\ul 47.43}    & \textbf{61.29}                    \\
VideoLLaVA-7B \citep{lin2023video}                        & 47.55          & 57.29          & 50.59          & 14.89          & 12.19          & 10.15          & 35.68          & 33.88          & 34.61          & 43.73                             \\
VideoChat2-7B \citep{2023videochat}                       & 43.20          & 48.98          & 43.00          & {\ul 30.85}    & 26.83          & {\ul 26.00}    & 26.74          & 28.92          & 33.33          & 39.11                             \\
ShareGPT4Video  \citep{chen2024sharegpt4video}                     & 51.48          & 62.24          & 48.27          & 17.02          & 19.51          & 8.91           & 43.00          & 38.84          & 32.05          & 46.28                             \\
LLaVA-v1.5-7B  \citep{liu2023llava}                              & 52.03          & 62.24          & 49.85          & 20.21          & 36.58          & 17.33          & 45.35          & 42.98          & 46.15          & 48.33                             \\
LLaMA-VID-7B   \citep{li2024llamavid}                      & 48.75          & 56.71          & 50.09          & 20.21          & 21.95          & 21.78          & 37.76          & 42.98          & 29.49          & 45.31                             \\
LLaMA-VID-13B   \citep{li2024llamavid}                     & 47.82          & 54.23          & 49.29          & 12.76          & 29.26          & 5.20           & 40.65          & 38.01          & 32.05          & 43.91                             \\
PLLaVA-7B    \citep{xu2024pllava}                        & 44.26          & 60.93          & 41.41          & 20.21          & 37.80          & 9.90           & 43.63          & 15.70          & 32.05          & 40.17                             \\
PLLaVA-13B   \citep{xu2024pllava}                         & {\ul 62.02}    & {\ul 69.39}    & {\ul 56.55}    & 21.28          & {\ul 50.00}    & 15.10          & 48.60          & 46.28          & 44.87          & 54.87                             \\
Qwen2.5-VL-3B-Instruct    \citep{qwen2.5-VL}           & 52.36          & 65.31          & 49.12          & 20.21          & 37.80          & 5.20           & 43.63          & 15.70          & 32.05          & 46.85                             \\
Qwen2.5-VL-7B-Instruct   \citep{qwen2.5-VL}             & 55.97          & 60.05          & 50.25          & 20.21          & 32.93          & 8.66           & 53.39          & 44.63          & 34.61          & 50.19                             \\
LLaVA-NeXT-Video-DPO-7B  \citep{zhang2024llavanext-video}            & 49.35          & 57.58          & 49.86          & 15.96          & 31.71          & 12.62          & 38.12          & 43.80          & 41.03          & 45.25                             \\
LLaVA-NeXT-Video-DPO-34B   \citep{zhang2024llavanext-video}           & 52.59          & 60.79          & 47.42          & 22.34          & 34.15          & 7.67           & 45.17          & 41.32          & 26.92          & 46.81                             \\
Video-LLaMA-2-13B    \citep{cheng2024videollama}                & 28.95          & 40.96          & 30.80          & 11.70          & 9.76           & 12.62          & 16.71          & 21.49          & 26.92          & 26.97                             \\
GPT-4o-mini    \citep{hurst2024gpt}                      & 52.87          & 59.47          & 54.79          & \textbf{58.51} & \textbf{64.63} & \textbf{62.13} & \textbf{63.50} & {\ul 52.89}    & \textbf{60.26} & {\ul 56.80}                       \\ 
\bottomrule
\end{tabular}
}
\caption{\textbf{Hallucination Evaluation.} Accuracy of LMMs on questions of different types of hallucination targets across three causes. The \textbf{bold} font indicates the best performance, and the {\ul underline} mark indicates the second-best.}
\label{hallucination}
\vspace{-0.4cm}
\end{table*}

\subsection{Evaluation Metrics}
\paragraph{Hallucination Evaluation.} To evaluate hallucinations in LMMs, we measure the \emph{accuracy} of their responses through LLM judging. Specifically, we employ GPT4o-mini~\cite{hurst2024gpt} as the evaluation model because it is widely regarded as having human-like judgment capabilities. However, since some models may exhibit extended reasoning processes, an optional GPT4o-mini-based extraction step is performed to isolate the model's answer before evaluation. Details can be found in Appendix~\ref{eval}.

\vspace{-0.4cm}
\paragraph{Consistency Evaluation.} 
To quantify response consistency in LMMs, we introduce \emph{bias score} to identify instances where a model provides inconsistent answers across different variants of the same base question. For binary-choice questions, the bias score is determined by the proportion of question pairs in which the model's responses differ between the "yes"-correct and "no"-correct versions. Similarly, for multiple-choice questions, bias is measured by the frequency of inconsistent responses among the three variant forms of each base question. A lower bias score indicates more consistent reasoning across question variants.

\section{Analysis of Video Hallucination}
\label{sec:anaysis}



\subsection{Settings}
We evaluated 16 LMMs in total: one 3B model, ten 7B models, two 13B models, one 34B model, and one API model. To ensure a fair comparison, we maintained the original settings for all baselines, including frame counts and generation hyperparameters. Details of these LMMs can be found in Appendix~\ref{imple}.

\subsection{Hallucination Evaluation}
Table~\ref{hallucination} demonstrates a performance comparison of various LMMs in hallucination evaluation across different dimensions with accuracy. Higher scores represent lower levels of hallucination. The evaluation results are categorized into three causes (prior conflict, in-context conflict, and capability deficiencies), with each category assessing three video aspects (object, scene, and event). Notably, Valley-Eagle-7B and GPT4o-mini emerged as the top performers, achieving total accuracy of 61.29\% and 56.80\% respectively. Valley-Eagle-7B demonstrated superior performance across all subtypes of prior knowledge conflicts, while GPT4o-mini excelled in all in-context conflict scenarios. In contrast, Video-LLaMA-2-13B and VideoChatGPT-7B showed the weakest performance, particularly struggling with in-context conflict situations. We present some failure cases in Appendix~\ref{subsec:LMMcase}.

\begin{figure*}[t!]
\centering
    \centering
    \begin{subfigure}[b]{0.328\textwidth}
        \centering
        \includegraphics[width=\textwidth]{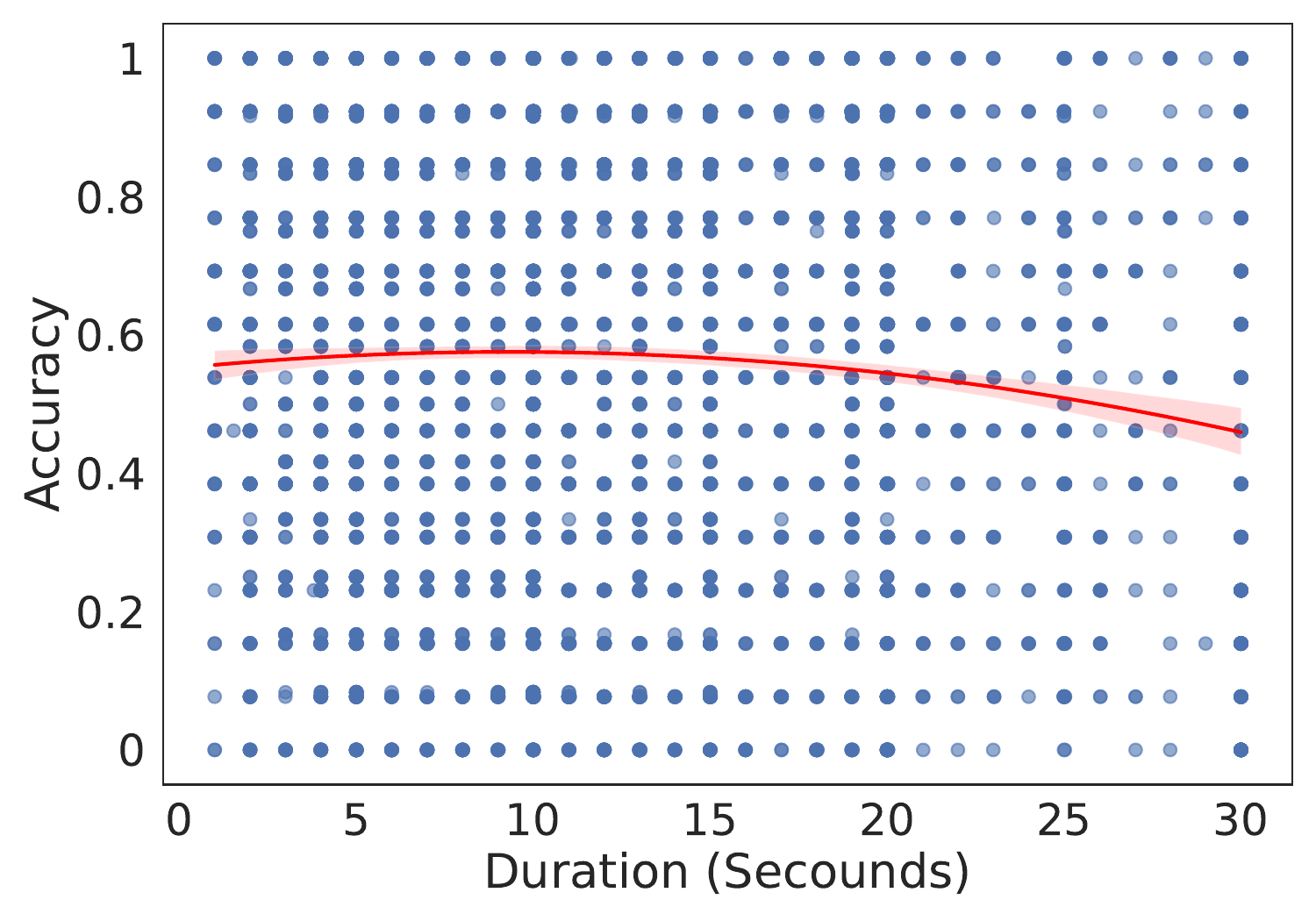}
        \caption{Duration Time}
        \label{hit:subfig1}
        \vspace{-0.2cm}
    \end{subfigure}
    \hfill
    \begin{subfigure}[b]{0.328\textwidth}
        \centering
        \includegraphics[width=\textwidth]{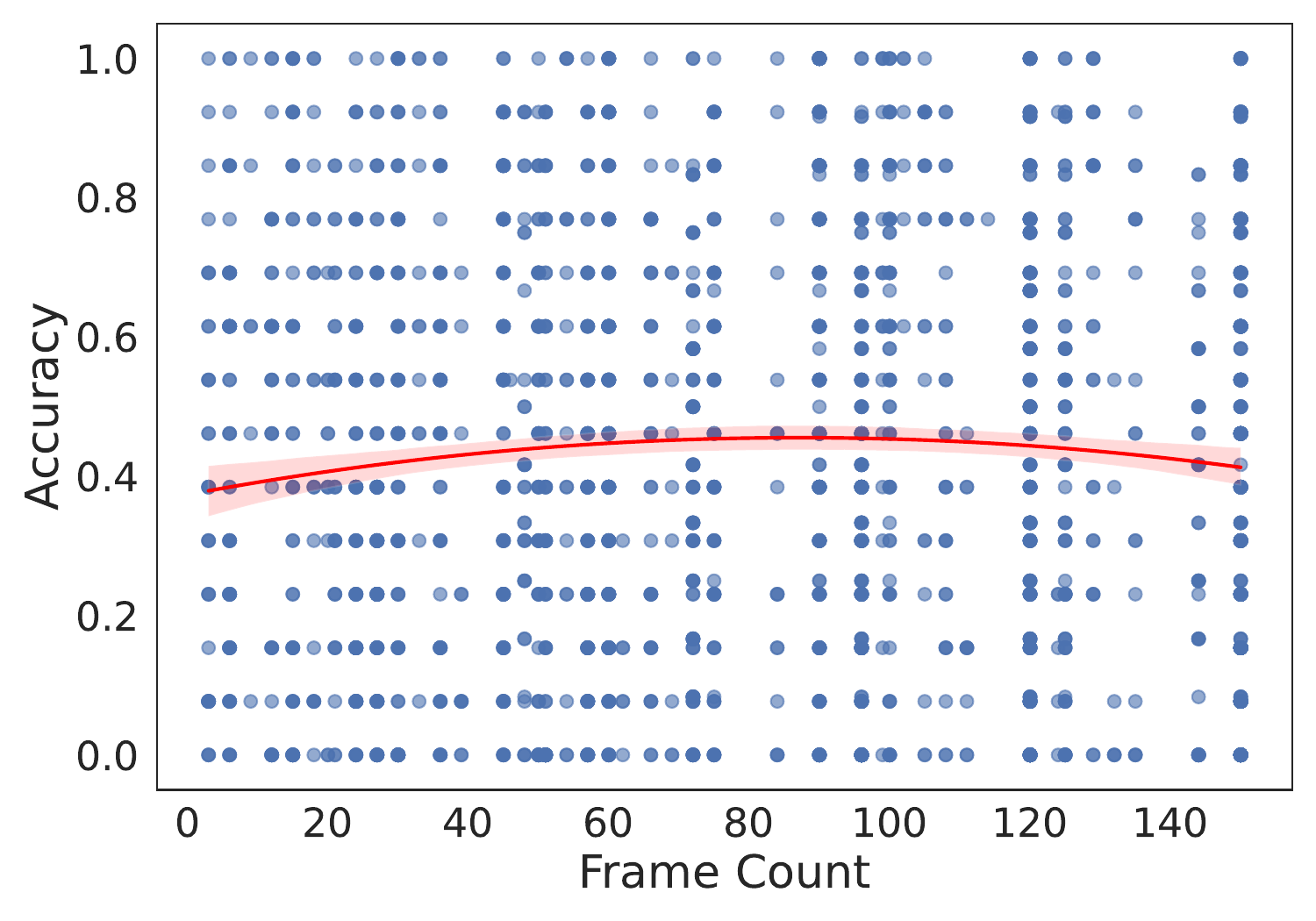}
        \caption{Frame Count}
        \label{hit:subfig2}
        \vspace{-0.2cm}
    \end{subfigure}
    \hfill
    \begin{subfigure}[b]{0.328\textwidth}
        \centering
        \includegraphics[width=\textwidth]{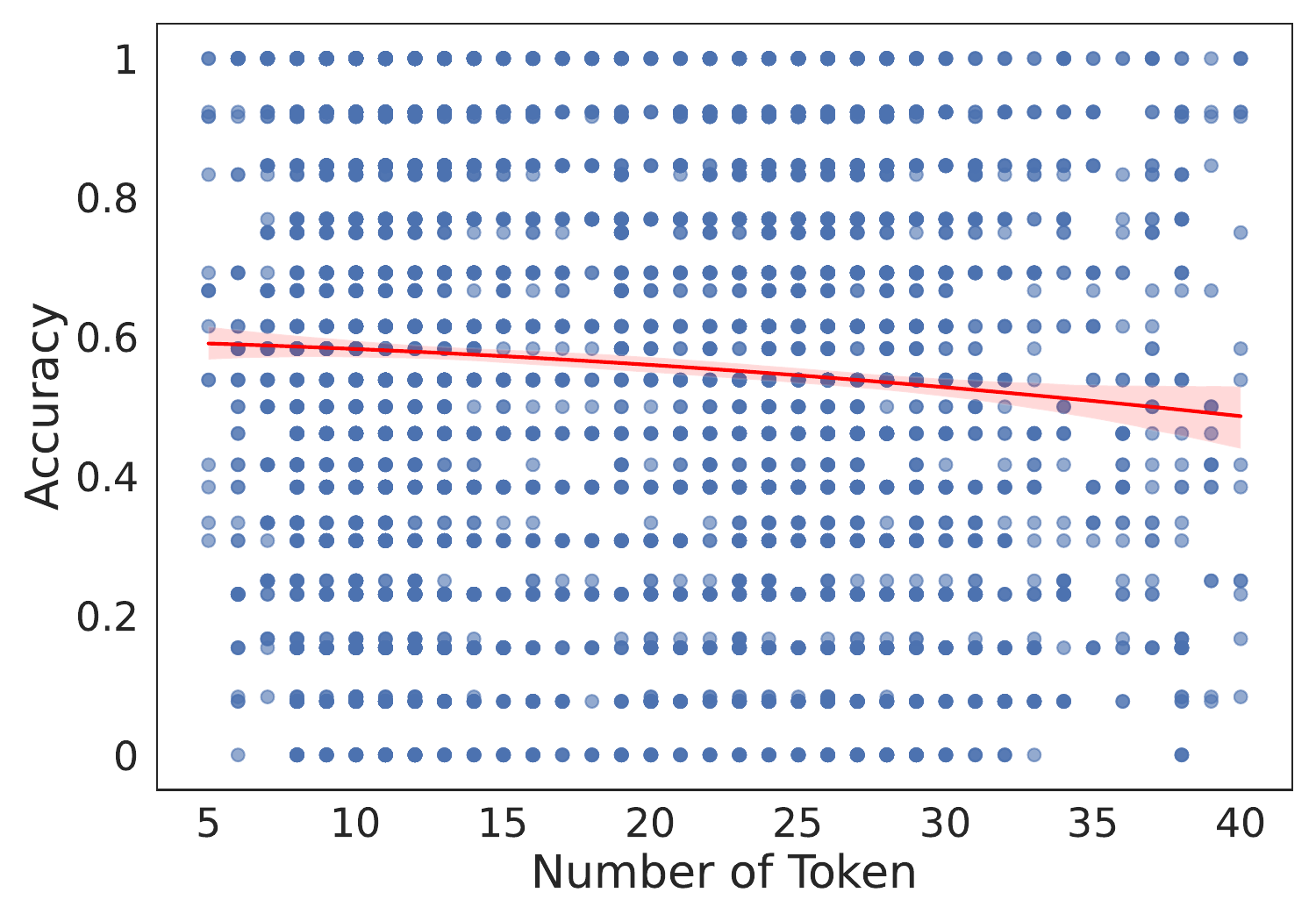}
        \caption{Question Length}
        \label{hit:subfig3}
        \vspace{-0.1cm}
    \end{subfigure}
    \caption{The impact of video duration, frame count, and question length on LLM hallucination.}
    \label{fig:impact}
    \vspace{-0.3cm}
\end{figure*}

\begin{figure*}[h]
\centering
    \centering
    \begin{subfigure}[b]{0.32\textwidth}
        \centering
        \includegraphics[width=\textwidth]{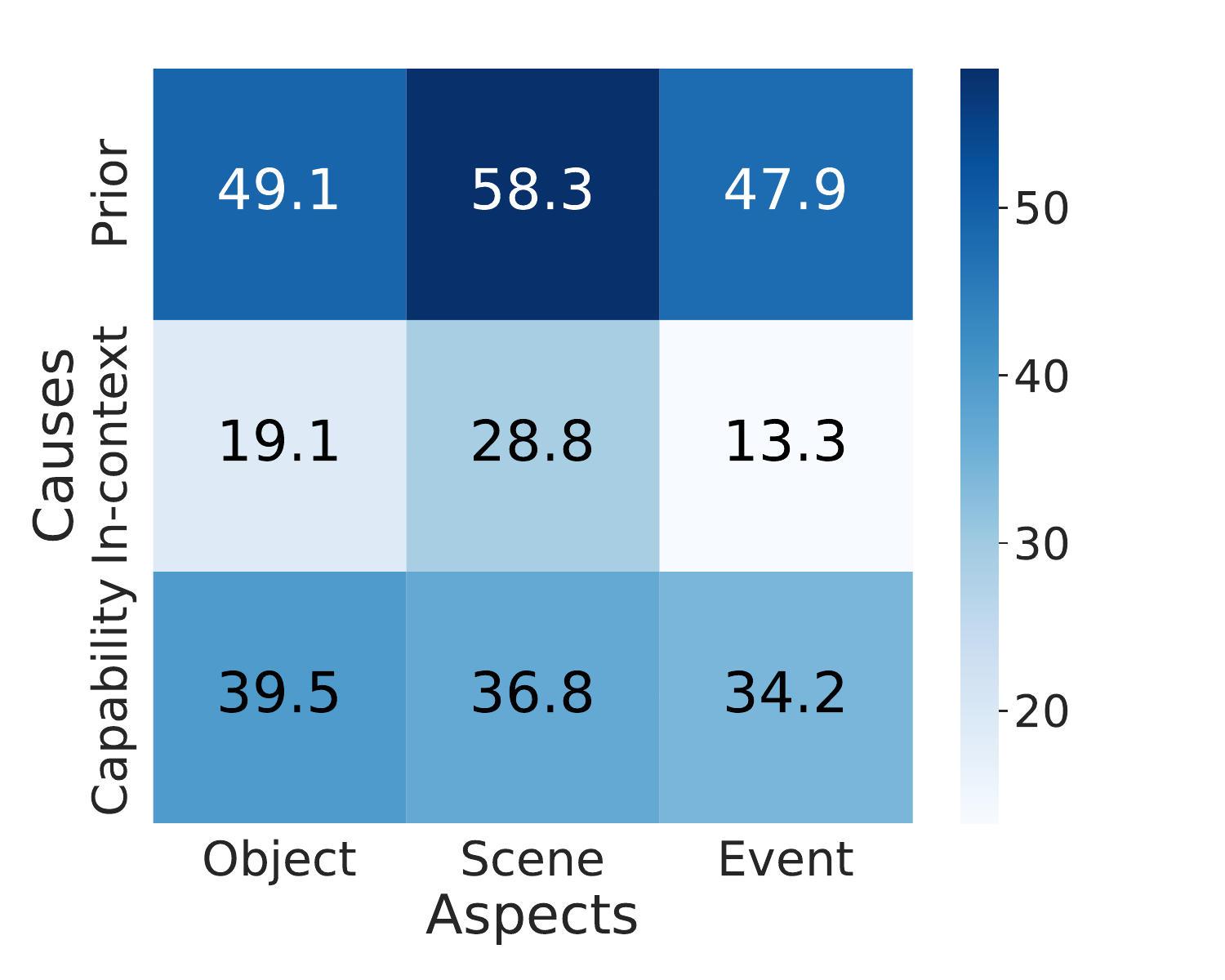}
        \caption{Causes-Aspects}
        \label{hit:subfig1}
    \end{subfigure}
    \hfill
    \begin{subfigure}[b]{0.32\textwidth}
        \centering
        \includegraphics[width=\textwidth]{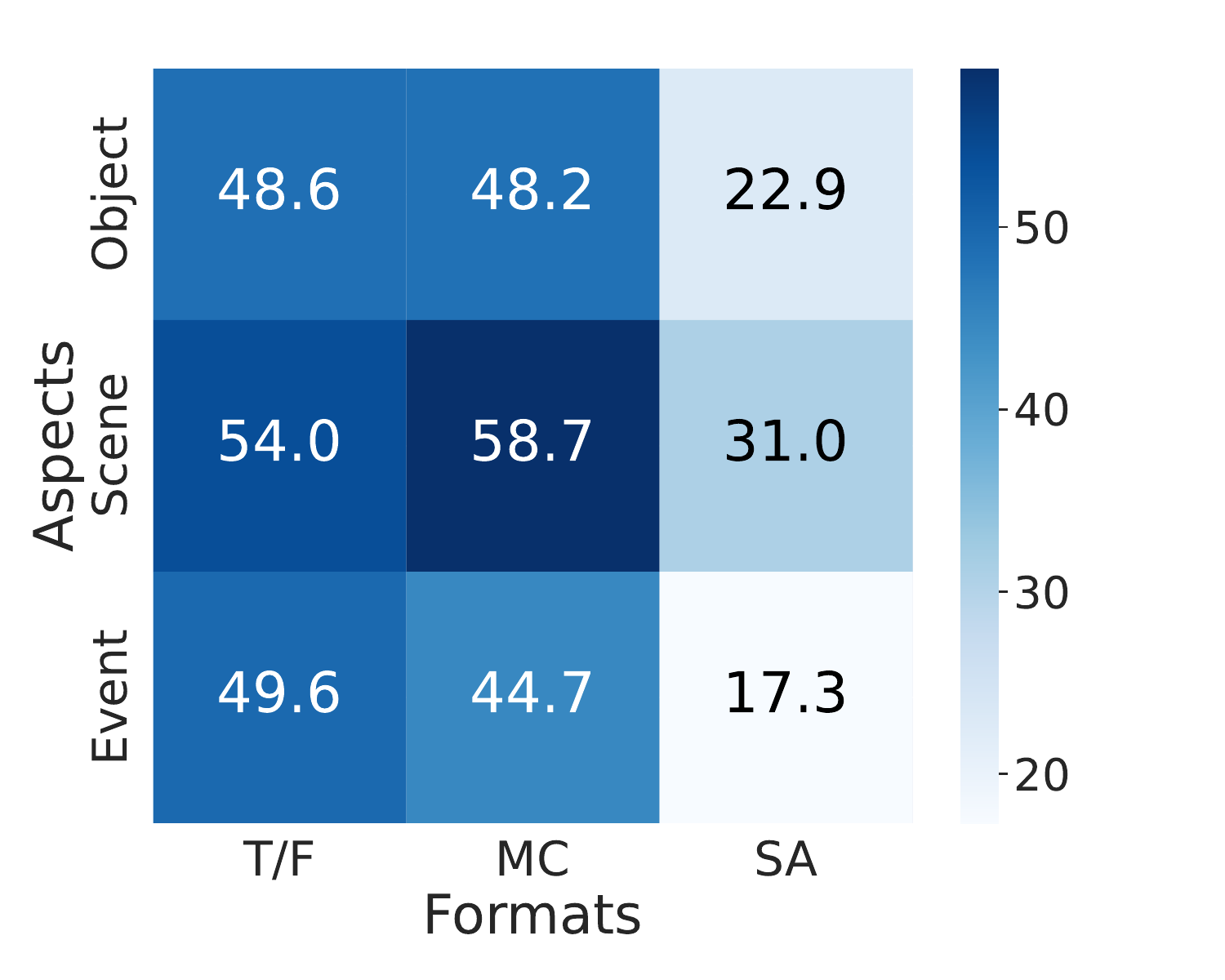}
        \caption{Formats-Aspects}
        \label{hit:subfig2}
    \end{subfigure}
    \hfill
    \begin{subfigure}[b]{0.32\textwidth}
        \centering
        \includegraphics[width=\textwidth]{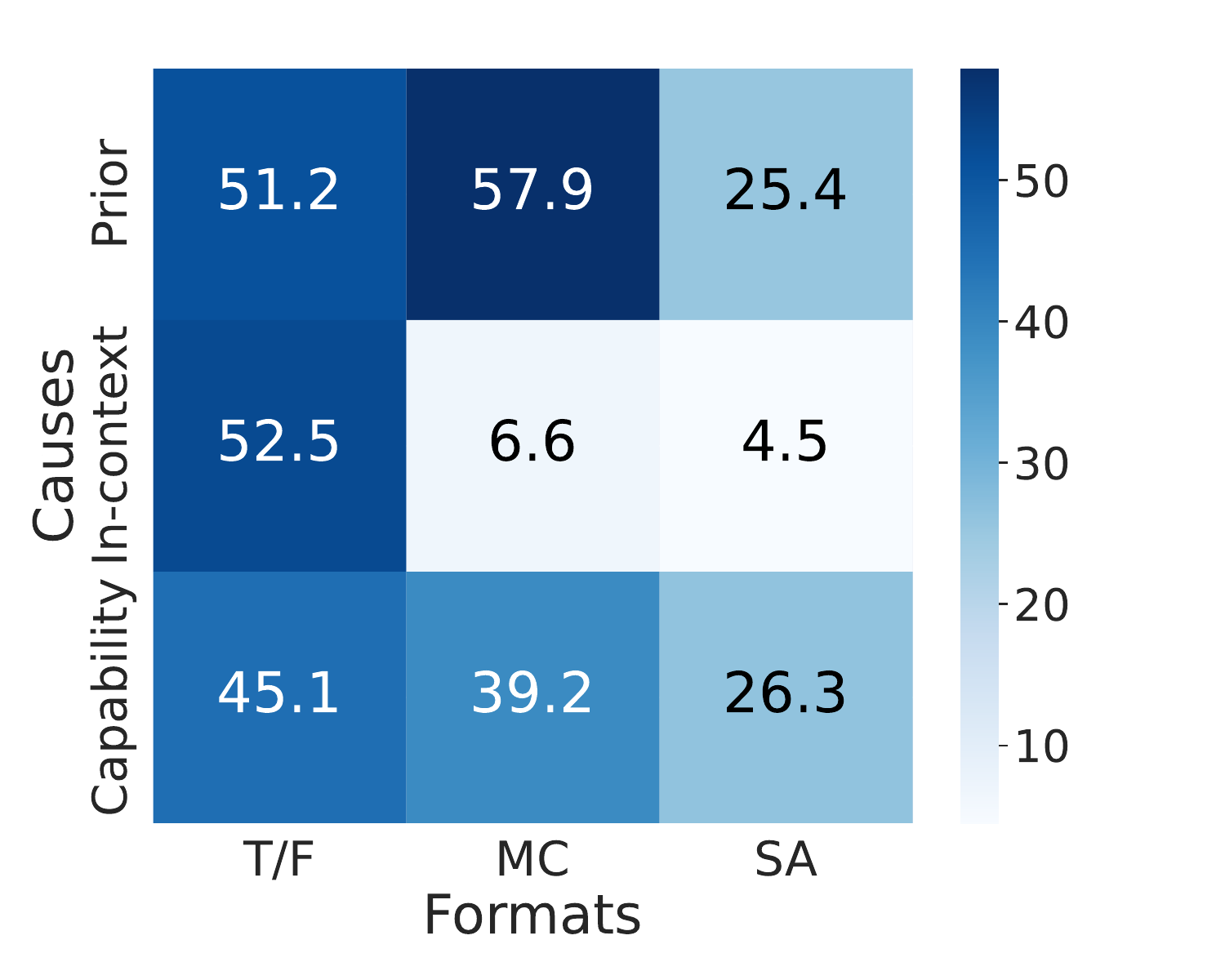}
        \caption{Formats-Causes}
        \label{hit:subfig3}
    \end{subfigure}
    \vspace{-0.2cm}
    \caption{Accuracy heatmap along two dimensions.}
    \vspace{-0.3cm}
    \label{fig:whole1}
\end{figure*}
\subsection{Consistency Evaluation}
Table~\ref{consis-} shows consistency evaluation with bias scores, where lower values indicate better performance. Qwen-VL-3B achieved the best overall score at 27.66\%, with Valley-Eagle-7B coming in second at 28.31\%. GPT-4o-mini demonstrated superior performance in binary-choice tasks with the lowest average bias score of 34.08\%, while Qwen-VL-3B excelled in multiple-choice scenarios with a score of 17.39\%. LLaVA-v1.5-7B also exhibited noteworthy consistency while maintaining low hallucination rates. In contrast, VideoChat2-7B and PLLaVA-7B showed the weakest performance in consistency evaluations.

\begin{table}[]
\centering
\begin{adjustbox}{width=0.47\textwidth}
\begin{tabular}{c|ccc}
\toprule
\textbf{Model/Type}  & \textbf{Binary-choice} & \textbf{Multiple-choice} & \textbf{Total} \\
\midrule
VideoChatGPT-7B      & 57.58                  & 39.19                    & 46.97         \\
Valley-Eagle-7B      & 42.41                  & {\ul 17.99}              & {\ul 28.31}   \\
VideoLLaVA-7B        & 59.08                  & 43.62                    & 50.16         \\
VideoChat2-7B        & 52.43                  & 49.75                    & 50.88         \\
ShareGPT4Video       & 60.11                  & 20.90                    & 37.48         \\
LLaVA-v1.5-7B        & 55.90                  & 35.38                    & 44.06         \\
LLaMA-VID-7B         & 52.15                  & 40.20                    & 45.25         \\
LLaMA-VID-13B        & 59.83                  & 30.75                    & 43.05         \\
PLLaVA-7B            & 61.14                  & 42.51                    & 50.38         \\
PLLaVA-13B           & 52.62                  & 24.02                    & 36.11         \\
Qwen2.5-VL-3B-Instruct        & {\ul 41.67}            & \textbf{17.39}           & \textbf{27.66} \\
Qwen2.5-VL-7B-Instruct        & 48.50                  & 19.90                    & 31.99         \\
LLaVA-NeXT-Video-DPO-7B  & 51.13                  & 41.41                    & 45.52         \\
LLaVA-NeXT-Video-DPO-34B & 49.06                  & 18.90                    & 31.65         \\
Video-LLaMA-2-13B    & 66.48                  & 24.32                    & 42.14         \\
GPT-4o-mini          & \textbf{34.08}         & 29.95                    & 31.69         \\
\bottomrule
\end{tabular}
\end{adjustbox}
\vspace{-0.2cm}
\caption{Consistence Evaluation}
\label{consis-}
\vspace{-0.6cm}
\end{table}

\subsection{On Length of Videos and Questions}
To analyze the relationship between LMM hallucination and various factors of input data, we conducted an analysis examining how accuracy varies with respect to video duration, frame count, and question length. As illustrated in Fig.~\ref{fig:impact}, an increase in video duration or frame count initially enhances performance, but beyond a certain point, the performance declines. This trend may be attributed to the fact that longer videos provide additional information; however, if the video is too long, the model may fail to sample the most relevant frames, leading to information loss. Conversely, as the length of the question text increases, the model’s performance consistently deteriorates, possibly because the surplus textual information diverts the model's attention away from the video content.





\subsection{On Sampling Number of Frames}
\begin{figure}[tt]
\begin{center}
\includegraphics[width=\linewidth]{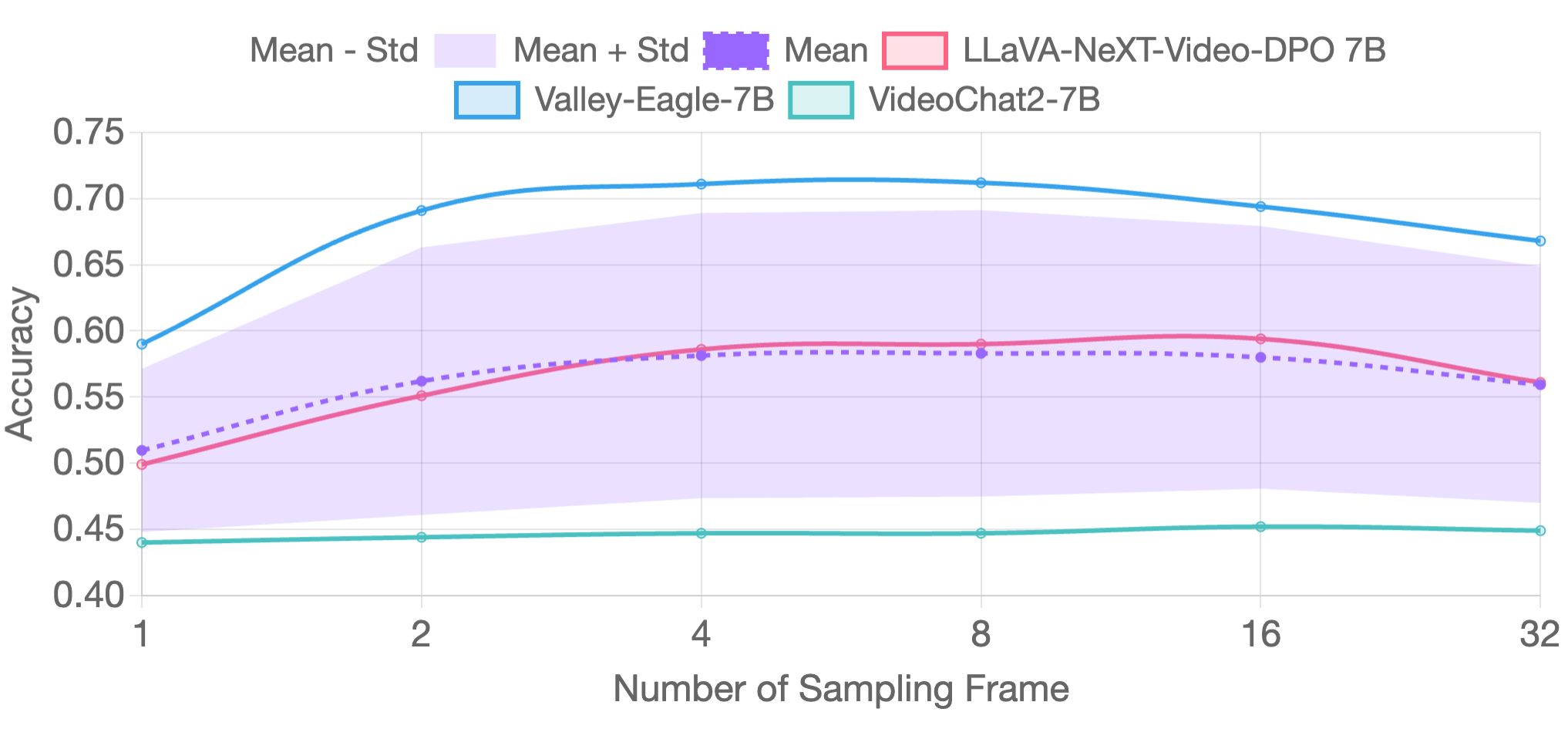}
\end{center}
\vspace{-0.6cm}
   \caption{Relationship between number of sampling frames and model performance.} 
    \label{frame}
     \vspace{-0.4cm}
\end{figure}

As video information is frame-dependent, we examine the relationship between the number of sampled frames and model performance. We choose three models to demonstrate overall trends with mean values and variances. Fig.~\ref{frame} shows how model hallucinations vary with increasing frame count. Performance improves with additional frames initially but deteriorates beyond a certain threshold, likely due to the growing disparity between training and inference frame counts.

\subsection{On Model Size}
\begin{figure}[t!]
\centering
    \centering
    \begin{subfigure}[b]{0.235\textwidth}
        \centering
        \includegraphics[width=\textwidth]{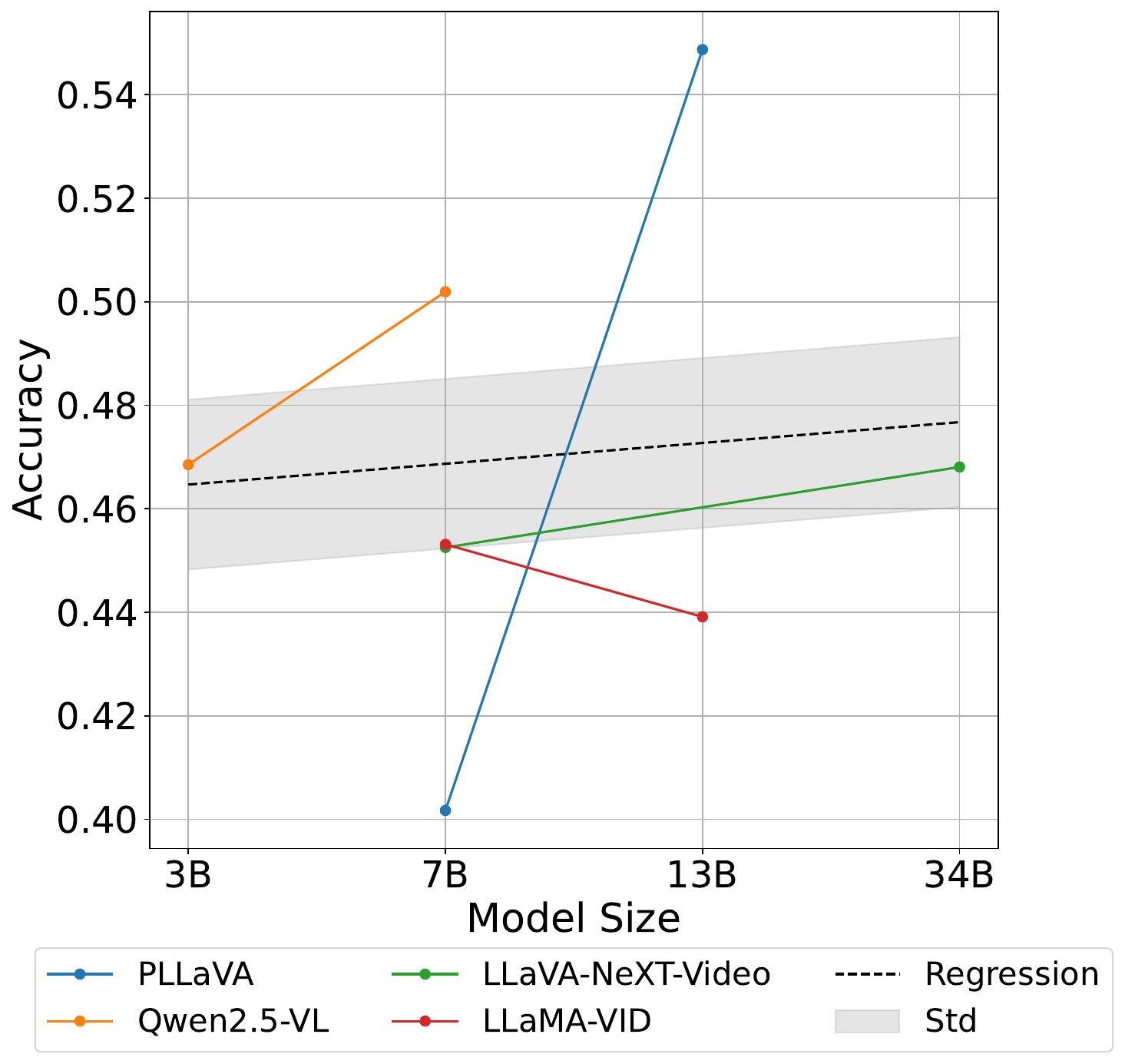}
        \caption{Hallucination}
        \label{hit:subfig1}
    \end{subfigure}
    \hfill
    \begin{subfigure}[b]{0.235\textwidth}
        \centering
        \includegraphics[width=\textwidth]{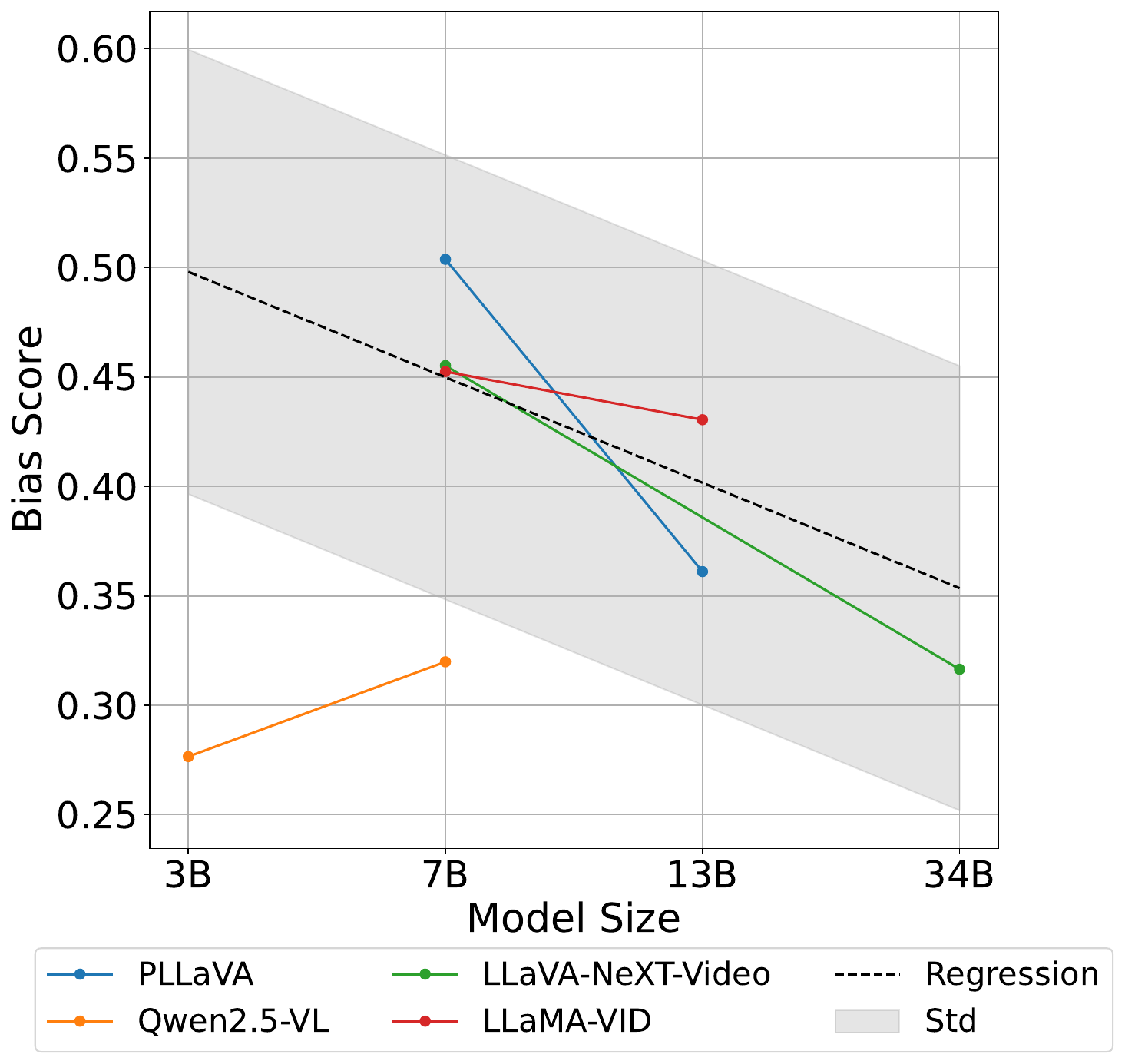}
        \caption{Consistency}
        \label{hit:subfig2}
    \end{subfigure}
    \vspace{-0.5cm}
    \caption{Relationship between model size and performance.}
    \label{fig:two11}
     \vspace{-0.6cm}
\end{figure}

Fig.~\ref{fig:two11} illustrates how accuracy and bias scores change as the number of parameters increases across different model categories. We selected four distinct types of model across various sizes and presented the regression lines. Although a few LMMs don't align perfectly with this trend, the overall evidence indicates that larger models tend to experience fewer hallucinations and exhibit lower levels of bias.

\subsection{On Co-impact of Each Two Dimensions}
In Fig.~\ref{fig:whole1}, we visualize the co-impact of each pair of dimensions by calculating the average accuracy across all models for corresponding categories in each dimension pair. The results demonstrate that regardless of the classification method, hallucinations caused by in-context conflicts consistently show the most significant impact across all combinations. Additionally, short-answer questions consistently exhibit lower performance scores across all combinations.
\begin{figure*}[t]
\begin{center}
\includegraphics[width=\linewidth]{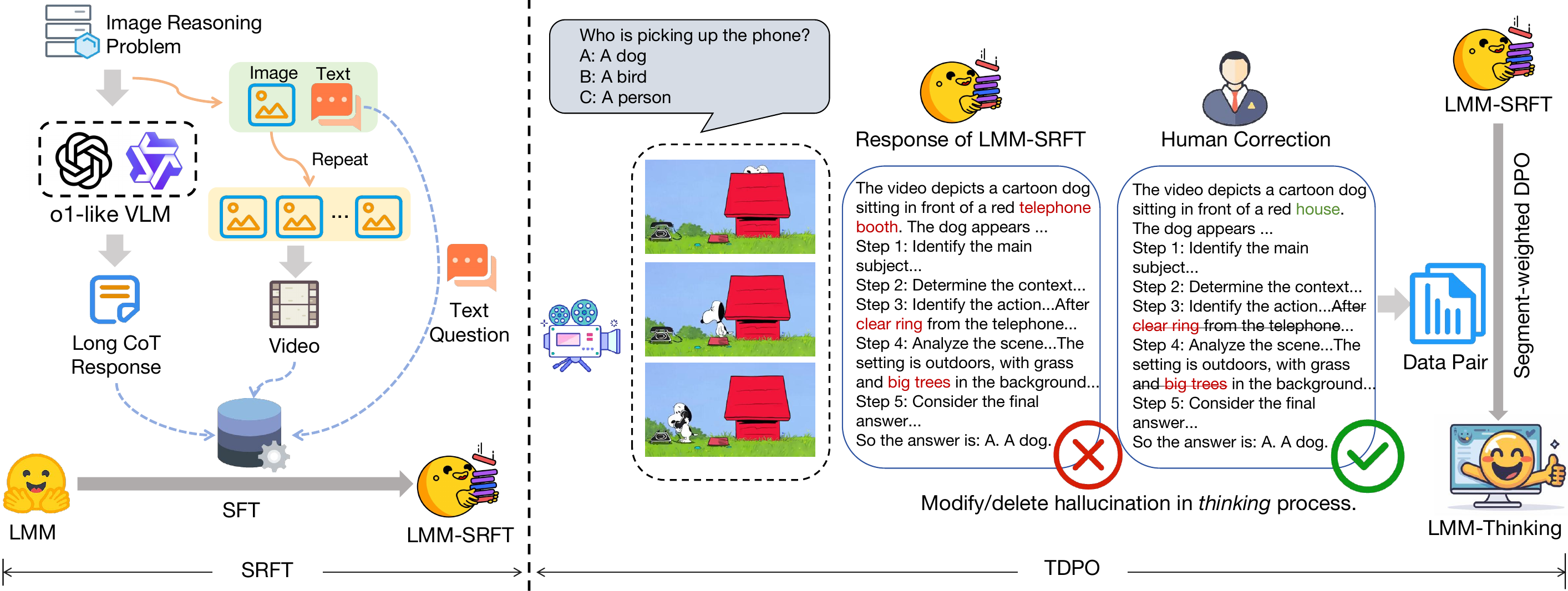}
\end{center}
\vspace{-0.3cm}
   \caption{\textbf{Our thinking-based training strategy.} Our training approach consists of two components: SRFT (for equipping the model with long reasoning capabilities) on the left, and TPDO (for preference learning to reduce hallucinations) on the right.} 
    \label{method}
\end{figure*}

\subsection{On Chain-of-Thought Reasoning}
\label{sec:cot}
To explore whether enhanced reasoning capabilities could reduce model hallucinations, we conducted a comparative study of three LMMs, evaluating their performance before and after implementing Chain-of-Thought~\cite{wei2022chain}(CoT). As shown in Fig.~\ref{tab:model-comparison-cot}, all LMMs demonstrated improved accuracy after incorporating CoT, indicating a reduction in hallucinations. Additionally, bias scores decreased for all LMMs except LLaMA-VID-7B, suggesting improved consistency.



\begin{table}[htbp]
    \centering
    \begin{adjustbox}{width=0.4\textwidth}
    \begin{tabular}{c|cc}
        \toprule
        \textbf{Model/Type} & \textbf{Accuracy $\uparrow$} & \textbf{Bias Score $\downarrow$} \\ 
        \midrule
        PLLaVA-7B & 40.17 & 50.38 \\
        +CoT      & 44.95 & 40.38 \\
        \midrule
        LLaMA-VID-7B & 45.31 & 45.25 \\
        +CoT         & 46.05 & 50.20 \\
        \midrule
        LLaVA-NEXT-Video-7B & 45.25 & 45.52 \\
        +CoT                 & 49.56 & 45.15 \\
        \bottomrule
    \end{tabular}
    \end{adjustbox}
    \caption{Performance comparison of the model before and after incorporating Chain-of-Thought (CoT)}
    \label{tab:model-comparison-cot}
    \vspace{-0.4cm}
\end{table}

\section{Mitigating Video Hallucination}

\label{sec:mitigation}

\begin{table*}[t!]
\centering
\resizebox{0.98\textwidth}{!}{
\begin{tabular}{c|cccccccccc}
\toprule
\multirow{2}{*}{\textbf{Model/Type}} & \multicolumn{3}{c}{\textbf{Prior}} & \multicolumn{3}{c}{\textbf{In-context}} & \multicolumn{3}{c}{\textbf{Capability}} & \multirow{2}{*}{\textbf{Total}} \\
\cmidrule(lr){2-4} \cmidrule(lr){5-7} \cmidrule(lr){8-10}
                                     & Object & Scene & Event & Object & Scene & Event & Object & Scene & Event &  \\ 
\midrule
PLLaVA                          & 44.26 & 60.93 & 41.41 & 20.21 & 37.80 & 9.90  & 43.63 & 15.70 & 32.05 & 40.17 \\
+CoT              & 49.44 & 53.50 & 44.13 & 35.10 & 58.54 & 38.37 & 36.77 & 33.88 & 29.49 & 44.95 \\ 
\midrule
LLaMA-VID-7B                    & 48.75 & 56.71 & 50.09 & 20.21 & 21.95 & 21.78 & 37.76 & 42.98 & 29.49 & 45.31 \\
+CoT                           & 50.04 & 58.45 & 50.54 & 24.47 & 25.61 & 26.98  & 35.50 & 39.67 & 30.77 & 46.05 \\ 
\midrule
LLaVA-NeXT-Video-DPO-7B              & 51.48 & 62.24 & 48.27 & 17.02 & 19.51 & 8.91  & 43.00 & 38.84 & 32.05 & 45.25 \\
+CoT                               & 52.08 & 64.28 & 50.88 & 36.17 & 51.22 & 34.41 & 41.01 & 47.93 & 37.18 & 49.56 \\ 
\midrule
LLaVA-NeXT-Video-DPO-34B             & 52.59 & 60.79 & 47.42 & 22.34 & 34.15 & 7.67  & 45.17 & 41.32 & 26.92 & 46.81 \\
GPT-4o-mini                      & 52.87 & 59.47 & 54.79 & 58.51 & 64.63 & 62.13 & 63.50 & 52.89 & 60.26 & 56.80 \\ 
\midrule
\rowcolor{gray!20}
LLaVA-NeXT-Video-7B-Thinking     & 58.28 & 69.97 & 53.37 & 39.36 & 56.10 & 40.84 & 39.38 & 37.19 & 34.61 & 52.90 \\ 
\bottomrule
\end{tabular}
}
\vspace{-0.1cm}
\caption{Hallucination evaluation of LLaVA-NeXT-Video-7B-Thinking comparing with other LMMs with/without CoT.}
\label{hallucination_think}
\vspace{-0.2cm}
\end{table*}


\begin{table}[t!]
    \centering
    \begin{adjustbox}{width=0.465\textwidth}
    \begin{tabular}{c|ccc}
    \toprule
    \textbf{Model/Type} & \textbf{Binary-choice} & \textbf{Multiple-choice} & \textbf{Total} \\ 
    \midrule
    PLLaVA & 61.14 & 42.51 &50.38  \\
    +CoT & 48.32 & 34.56 &40.38  \\ 
    \midrule
    LLaMA-VID-7B & 52.15 & 40.20 &45.25  \\
    +CoT & 65.70 & 38.86 &50.20  \\ 
    \midrule
    LLaVA-NeXT-Video-DPO-7B & 51.13 & 41.41 &45.52  \\
    +CoT & 56.84 & 36.58 &45.15  \\ 
    \midrule
    LLaVA-NeXT-Video-DPO-34B & 49.06 &  18.90 & 31.65 \\
    GPT-4o-mini & 34.08 & 29.95 &31.69  \\ 
    \midrule
    \rowcolor{gray!20}
    LLaVA-NeXT-Video-7B-Thinking & 49.16 & 35.07 & 41.02 \\ 
    \bottomrule
    \end{tabular}
    \end{adjustbox}
    \vspace{-0.1cm}
    \caption{Consistence evaluation of LLaVA-NeXT-Video-7B-Thinking comparing with other LMMs with/without CoT.}
    \label{coni}
    \vspace{-0.2cm}
\end{table}


The o1-like models~\cite{jaech2024openai,guo2025deepseek,team2025kimi} have demonstrated that long chain-of-thought (CoT) reasoning can significantly improve the capability and reliability of large models due to their thinking progress. Inspired by these o1-like models and the observation in Sec.~\ref{sec:cot}, we apply supervised reasoning fine-tuning (SRFT) along with thinking-based direct preference optimization(TDPO) on our large video model, ultimately developing a video-thinking LMM that effectively mitigates video hallucination. The overview of our method is illustrated in Fig.~\ref{method}.

\subsection{Supervised Reasoning Fine-tuning}

To unlock the reasoning ability of LMMs, we proposed supervised reasoning 
 fine-tuning (SRFT) on base model $\mathcal{M}_{\text{base}}$. This method consists two part: reasoning data synthesis and LoRA-based SFT.
\vspace{-0.2cm}
\paragraph{Reasoning Data Synthesis.} 
Due to the limitation that existing o1-like models either lack support for visual input or are restricted to processing a single image at a time, it is not feasible to directly employ such models for generating video question-answer data. To address this limitation, we input both the question and image data into QVQ~\cite{qvq-72b-preview} to generate a reasoning response. Subsequently, by replicating the image to form a static video, we combine the question, the static video, and the generated reasoning response to create a training dataset. This dataset is designed to enhance the thinking capabilities of models with video understanding.
\vspace{-0.6cm}
\paragraph{LoRA-based SFT.}  
After constructing the reasoning training dataset \(\mathcal{D} = \{(v_i, x_i, y_i)\}_{i=1}^N\)—where \(v_i\) denotes the input video, \(x_i\) corresponds to text inputs, and \(y_i\) represents the target outputs—we perform supervised fine-tuning (SFT) with LoRA~\cite{hu2022lora} on the base model \(\mathcal{M}_{\text{base}}\) that employs a weight matrix \(W \in \mathbb{R}^{d \times k}\); since full-parameter fine-tuning may overwrite some prior knowledge and we only need to equip the model's ability to generate long CoT style, we instead adopt a low-rank adaptation strategy by introducing the update \(\Delta W = BA\) with \(B \in \mathbb{R}^{d \times r}\) and \(A \in \mathbb{R}^{r \times k}\) where \(r \ll \min(d, k)\), such that the adapted weight becomes \(W' = W + \alpha BA\) (with \(\alpha\) as a scaling factor).  It is formulated as follows:
\begin{equation}
\small
    L_{\text{SFT}}(A,B) \;=\; -\mathbb{E}_{(v,x,y) \sim \mathcal{D}} \log P_{\theta}\Bigl(y \,\Big|\, v,\, x;\; W + \alpha BA\Bigr)
\end{equation}
where $P_{\theta}$ denotes the model parameters. Through SRFT, we unlock the fundamental reasoning capabilities of the base model $M_{\text{base}}$, resulting in LMM-SRFT, which acquires the ability to generate long CoT reasoning. 
\subsection{Thinking-based DPO}
Although SRFT endows the LMM with reasoning capability, the thought process still exhibits hallucinations. Inspired by RLHF-V~\cite{yu2024rlhf}, fine-grained preference learning is well-suited for mitigating hallucinations in reasoning steps. This is because hallucinations in reasoning have two key characteristics: first, they manifest in the chain-of-thought as sporadic occurrences of specific words; and second, the thinking step naturally segments itself,
which makes it straightforward to assign varying preference weights to different segments based on their degree of hallucination. Therefore, we propose TDPO, a weighted DPO method tailored for the thinking process. Specifically, TDPO is divided into two stages: constructing the fine-grained preference data and applying segment-weighted DPO.
\vspace{-0.35cm}
\paragraph{Preference Data Construction.}
Given a video and a corresponding question, we initially employ the LMM to generate a response. Subsequently, a manual review is conducted to remove or modify any steps in the reasoning process that exhibit hallucinations. The original model-generated response is then treated as negative sample, while manually revised response serves as positive sample, thereby enabling the construction of the preference data.
\vspace{-0.4cm}
\paragraph{Segment-weighted DPO.}
Refer to DDPO in RLHF-V~\cite{yu2024rlhf}, we score the response as a weighted aggregation of the fine-grained thinking segments. Specifically, the likelihood  of generating an output response \( y \) given an input question \( x \) and video \( v \) is defined as:
\vspace{-0.2cm}
\begin{equation}
\small
\begin{split}
\log \pi(y|x,v) &= K\Biggl[ \sum_{y_i \in y_o} \log p\bigl(y_i \mid x,v, y_{<i}\bigr) \\[1mm]
    &\quad\quad + \gamma \sum_{y_i \in y_h} \log p\bigl(y_i \mid x,v, y_{<i}\bigr) \Biggr],
\end{split}
\vspace{-0.2cm}
\end{equation}
where \( y_o \) and \( y_h \) respectively denote the token sets corresponding to the original and human-corrected segments. Unlike the standard DPO~\cite{rafailov2023direct}---which uses the likelihood \(
\log \pi(y \mid x, v) = \sum_{y_i \in y} \log p\bigl(y_i \mid x, v, y_{<i}\bigr)\) as action score---treats all segments equally, our segment-weighted DPO assigns greater weight to the corrected reasoning steps, thereby reinforces the human preference for responses that are factually grounded.
Besides, the tradeoff hyperparameter \( \gamma \) can increase the emphasis on the corrected segments. As with DDPO, we also use a normalization constant,
\(K = 1/({|y_o| + \gamma |y_h|})\),
where \( |y_o| \) and \( |y_h| \) denote the lengths of the original and human corrected segments, respectively. This constant prevents longer responses from receiving disproportionately high scores.
By applying segment-weighted likelihood within the DPO using our preference dataset, we can train LMM-SRFT to exhibit LMM-Thinking.


\subsection{Experiment Result}
We conducted our thinking-based training on LLaVA-NeXT-Video-DPO-7B with about 5K synthetic data for SRFT and 3K preference data for TDPO. As shown in Table~\ref{hallucination_think}, the base model trained using our method—LLaVA-NeXT-Video-7B-Thinking—outperformed the original LLaVA-NeXT-Video-7B by 7.65\% and achieved a 3.34\% improvement over the original model with CoT. Notably, it has even surpassed the performance of the 34B-parameter LLaVA-NeXT-Video-DPO. Table~\ref{coni} shows the consistency evaluation of LLaVA-NeXT-Video-7B-Thinking alongside other LMMs. It demonstrates that LLaVA-NeXT-Video-7B-Thinking exhibits a higher consistency compared to the original LLaVA-NeXT-Video-DPO-7B—both with and without CoT—achieving bias score reductions of 4.5\% and 4.13\% respectively. We present some cases of response from LLaVA-NeXT-Video-7B-Thinking in Appendix~\ref{subsec:ourcase}.

\vspace{-0.1cm}
\section{Conclusion}
\vspace{-0.1cm}
\label{sec:conclusion}

In this study, we presented HAVEN, a comprehensive benchmark specifically designed for evaluating hallucinations in video understanding for LMMs. The benchmark is built around three key dimensions and consists of 6,497 questions. Extensive evaluation across 16 models revealed insights into how factors such as video and question length, frame sampling and model size collectively impact model hallucinations. Furthermore, we proposed the SRFT and TDPO strategies for training video-thinking models. After applying to LLaVA-NeXT-Video-DPO-7B, our approach both improved its accuracy in hallucination evaluation and reduced its bias score in consistency evaluation. We will include more open-source and API LMMs in the future while also exploring automatic correction strategies to generate TDPO data instead of relying on human corrections.




{
    \small
    \bibliographystyle{ieeenat_fullname}
    \bibliography{main}
}

\clearpage
\appendix
\newpage
\setcounter{page}{1}
\maketitlesupplementary
\section{Evaluation Details}
\label{eval}
Our evaluation is conducted using GPT-4o-mini~\cite{hurst2024gpt} by comparing human-generated answers with model responses to determine correctness. 

For models without CoT, responses can be directly compared using GPT-4o-mini. Multiple-choice questions are evaluated using the prompt in Table~\ref{tab:multiple_choice_prompt}, binary-choice questions using the prompt in Table~\ref{tab:binary_choice_prompt}, and short-answer questions using the prompt in Table~\ref{tab:short_answer_prompt}. 

For models using CoT, we found that long reasoning process might affect GPT-4o-mini's judgment in multiple-choice and binary-choice questions. Therefore, we first use GPT-4o-mini to extract a concise answer from the model’s reasoning process based on the given options in the question. Then, we can conduct our evaluation based on the extract answer. For multiple-choice questions, we let GPT-4o-mini directly assess whether the model’s response aligns with the intended meaning of the correct answer based on its reasoning process. Multiple-choice questions are evaluated using the prompt in Table~\ref{tab:Cot_BC_prompt}, binary-choice questions using the prompt in Table~\ref{tab:Cot_MC_prompt}, and short-answer questions using the prompt in Table~\ref{tab:Cot_SA_prompt}.

\section{Implementation Details of the LMMs}
\label{imple}

The models evaluated in our experiment are as follows:

\textbf{Video-ChatGPT} \citep{Maaz2023VideoChatGPT} combines the capabilities of LLMs with a pretrained visual encoder adapted for spatiotemporal video representation. The Video-ChatGPT model used in the experiment has a parameter size of 7B. For each video, we sample 100 frames and resize them to a resolution of 224×224.

\textbf{Valley-Eagle}  \citep{wu2025valley2} is a cutting-edge multimodal large model designed to handle a variety of tasks involving text, images, and video data, which is developed by ByteDance. The Valley-Eagle model used in the experiment has a parameter size of 7B. For each video, we sample 100 frames and resize them to a resolution of 384×384.

\textbf{Video-LLaVA} \citep{lin2023video} unifies visual representation into the language feature space to advance the foundational LLM towards a unified LVLM. The Video-LLaVA model used in the experiment has a parameter size of 7B. For each video, 8 frames are uniformly sampled from the video and resize to 224*224 definition as visual input. The normalized mean and standard deviation of video frames are set to (0.48145466, 0.4578275, 0.40821073) and (0.26862954, 0.26130258, 0.27577711), respectively.

\textbf{VideoChat2} \citep{2023videochat} integrates video foundation models and large language models via a learnable neural interface. The VideoChat2 model used in the experiment has a parameter size of 7B. For each video, it uniformly samples 8 frames from the video as visual input. The normalized mean and standard deviation of video frames are set to (0.48145466, 0.4578275, 0.40821073) and (0.26862954, 0.26130258, 0.27577711), respectively.

\textbf{ShareGPT4Video} \citep{chen2024sharegpt4video} facilitates the video understanding of large video-language models and the video generation of text-to-video models via dense and precise captions. The parameter size of the ShareGPT4Video model used in the experiment is 8B. For each video, it samples 16 frames as the visual input.

\textbf{LLaVA} \citep{liu2023llava} is an end-to-end trained large multimodal model that connects a vision encoder and LLM for general-purpose visual and language. The parameter size of the ShareGPT4Video model used in the experiment is 7B. It randomly samples a frame and resizes it to 336×336 as the visual input.

\textbf{LLaMA-VID} \citep{li2024llamavid} use a dual-token strategy to significantly reduce the overload of long videos while preserving critical information. In the experiments of this paper, two parameter sizes of the LLaMA-VID model were evaluated, including 7B and 13B. For each second of the video, the model samples one frame as visual input.

\textbf{PLLaVA} \citep{xu2024pllava} proposes a simple but effective pooling strategy to smooth the feature distribution along the temporal dimension and thus reduce the dominant impacts from the extreme features. In the experiments of this paper, two parameter sizes of the PLLaVA model were evaluated, including 7B and 13B. For each video, it uniformly samples 16 frames from the video and resize them to a resolution of 336×336 as visual input. The  target pooling shape is set to be 16 × 12 × 12 × d, where d corresponds to the input dimension of the LLMs.

\textbf{Qwen2.5-VL-Instruct} \citep{qwen2.5-VL} introduces dynamic resolution processing and absolute time encoding, enabling it to process images of varying sizes and videos of extended durations with second-level event localization. In the experiments of this paper, two parameter sizes of the Qwen2.5-VL-Instruct model were evaluated, including 3B and 7B. For each second of the video, the model samples two frames as visual input.

\begin{table*}[!]
\vspace{-0.4cm}
\centering
\begin{tikzpicture}
    \node[draw, fill=gray!10, rounded corners, inner sep=10pt] (box) {
        \begin{minipage}{0.95\textwidth}
            \textbf{\texttt{Multiple-Choice:}} \\
            
            You are a professional homework grading tool. I will provide you with four rules for grading: \\
            1. This is a multiple-choice question. Judge the correctness based on the selected letter and actual content of the provided answers.  \\ 
            2. Regardless of the question type, respond only with either 1 or 0, without any additional explanation.    \\
            3. 1 means the prediction is correct, and 0 means it is incorrect.   \\
            4. If the predicted answer matches the correct answer in meaning, even if it is phrased differently, consider it correct.   \\
            For example, if the prediction conveys the same meaning as the standard answer, you should respond with 1.  \\
            Based on the question and its standard answer, is the prediction correct? If yes, return only 1; otherwise, return only 0.  \\
            Question: \texttt{\{question[idx]\}}  \\
            Standard Answer: \texttt{\{answer[idx]\}}  \\
            The Predicted Answer: \texttt{\{extract\_answer\}}  \\
        \end{minipage}
    };
\end{tikzpicture}

\caption{The evaluation prompts for multiple-choice questions employed in the experiment for models without CoT.}
\label{tab:multiple_choice_prompt}
\end{table*}

\begin{table*}[!]
\centering
\begin{tikzpicture}
    \node[draw, fill=gray!10, rounded corners, inner sep=10pt] (box) {
        \begin{minipage}{0.95\textwidth}
            \textbf{\texttt{Binary-Choice:}} \\
            
            You are a professional homework grading tool. I will provide you with four rules for grading: \\
            1. This is a yes/no question. The Standard Answer is only Yes/No, please directly compare the standard answer with 'yes' or 'no' in the predicted answer. \\
            2. No matter what kind of questions, only response with one of 1 or 0. No more explanation.  \\
            3. 1 means correct (they are the same), 0 means wrong (they are different).  \\
            For example, if the prediction conveys the same meaning as the standard answer, you should respond with 1. \\
            Based on the question, is the prediction correct? If yes, only return 1, otherwise only return 0. \\
            Question: \texttt{\{question[idx]\}}  \\
            Standard Answer: \texttt{\{answer[idx]\}} \\
            The Predicted Answer: \texttt{\{extract\_answer\}} 
            \\
        \end{minipage}
    };
\end{tikzpicture}
\vspace{-0.1cm}
\caption{The evaluation prompts for Binary-Choice questions employed in the experiment for models without CoT.}
\label{tab:binary_choice_prompt}
\vspace{-0.4cm}
\end{table*}

\textbf{LLaVA-NEXT-Video-DPO} \citep{zhang2024llavanext-video} expands the applications to multi-image scenarios including multi-image, multi-frame (video), multi-view (3D), and multi-patch (single-image) scenarios. In the experiments of this paper, two parameter sizes of the LLaVA-NEXT-Video-DPO model were evaluated, including 7B and 34B. For each video, it uniformly samples 32 frames from video and resize them to a resolution of 336×336 as visual input.

\textbf{Video-LLaMA2} Video-LLaMA2 \citep{cheng2024videollama} incorporates a custom spatial temporal convolution (STC) connector, which effectively captures the intricate spatial and temporal dynamics of video data. For each video, it uniformly samples 8 frames from the video and resize them to a resolution of 336×336 as visual input.

\textbf{GPT-4o-mini} GPT-4o-mini \citep{hurst2024gpt} is a low-cost, low-latency online model launched by OpenAI, capable of handling multimodal tasks in various complex scenarios.  For each video, we sample 16 frames uniformly and resize to 768 pixels, keeping the original aspect ratio unchanged.

\vspace{-0.1cm}
\section{Case Demonstration}
\vspace{-0.1cm}
In addition to the experimental effects presented in Section {\color{cvprblue}4} and Section {\color{cvprblue}5.3}, we also present some response cases for different baseline LMMs and our improved model.

\subsection{Evaluated baseline LMMs}
\label{subsec:LMMcase}
\begin{figure*}[b]

\vspace{-0.45cm}
\begin{center}
\includegraphics[width=0.92\linewidth]{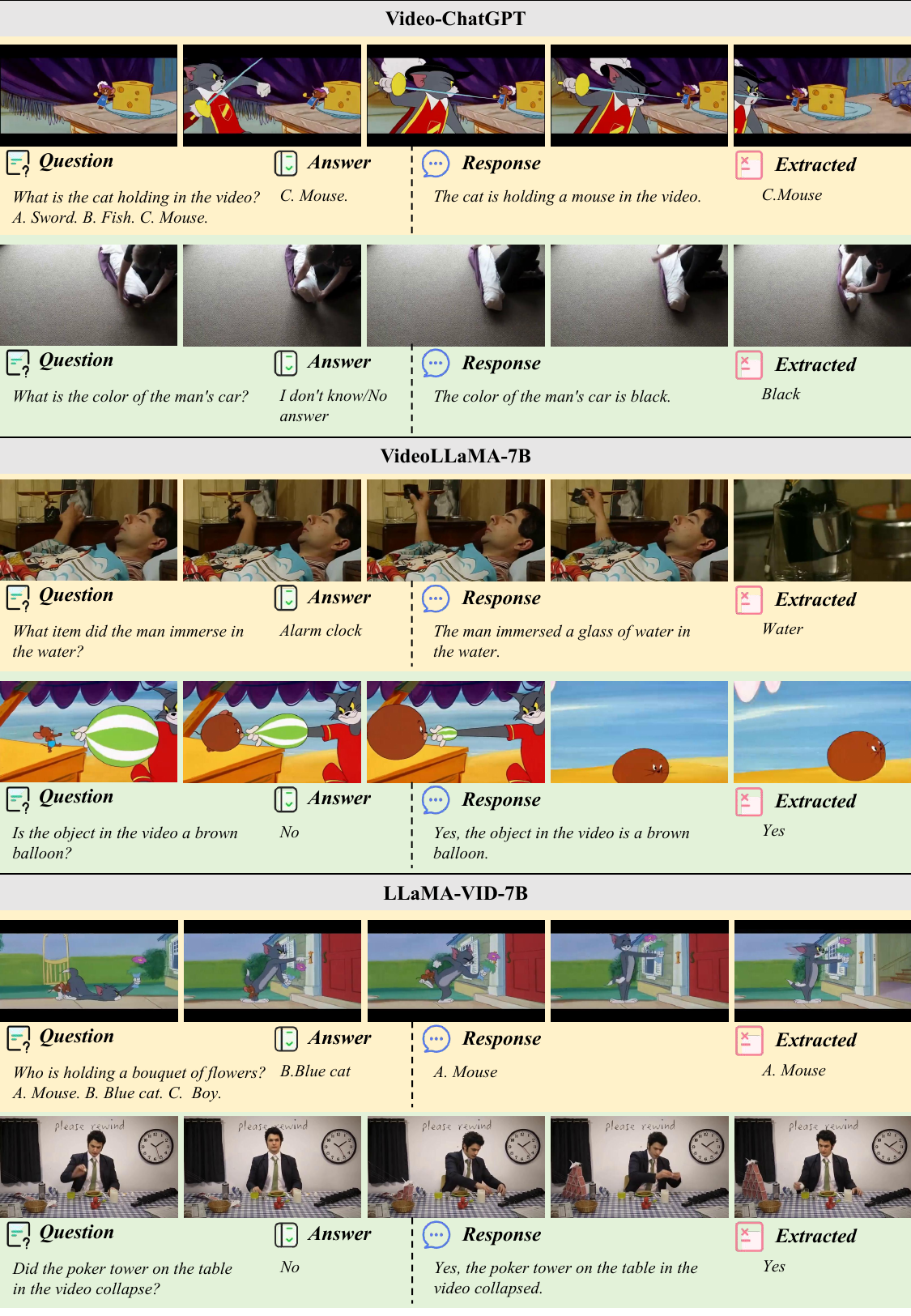}
\end{center}
\vspace{-0.52cm}
   \caption{Case of response from evaluated baseline LMMs. } 
    \label{case1}
\end{figure*}

\begin{figure*}[t]
\vspace{-0.45cm}
\begin{center}
\includegraphics[width=0.92\linewidth]{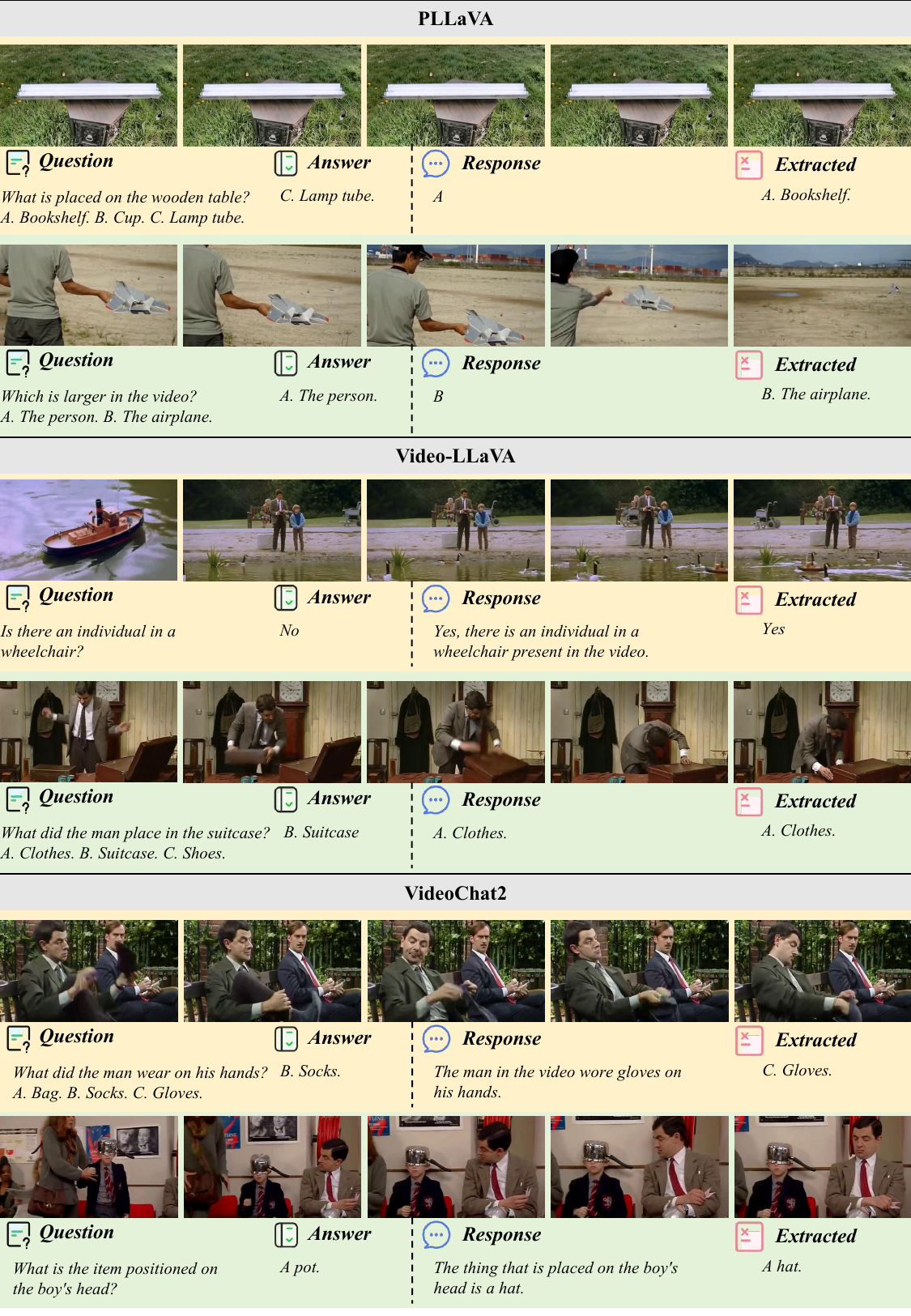}
\end{center}
\vspace{-0.52cm}
   \caption{Case of response from evaluated baseline LMMs. } 
    \label{case2}
\end{figure*}

Fig.\ref{case1} and Fig.\ref{case2} present examples of hallucinations observed in various LMMs. The responses from these models are concise and highly susceptible to hallucination. These examples highlight several factors contributing to the hallucinations analyzed above, including conflict with prior knowledge, conflict between contexts, and the capability deficiencies of those LMMs.
\subsection{LLaVA-NeXT-Video-7B-Thinking}
\label{subsec:ourcase}
\begin{figure*}[t]
\begin{center}
\includegraphics[width=0.92\linewidth]{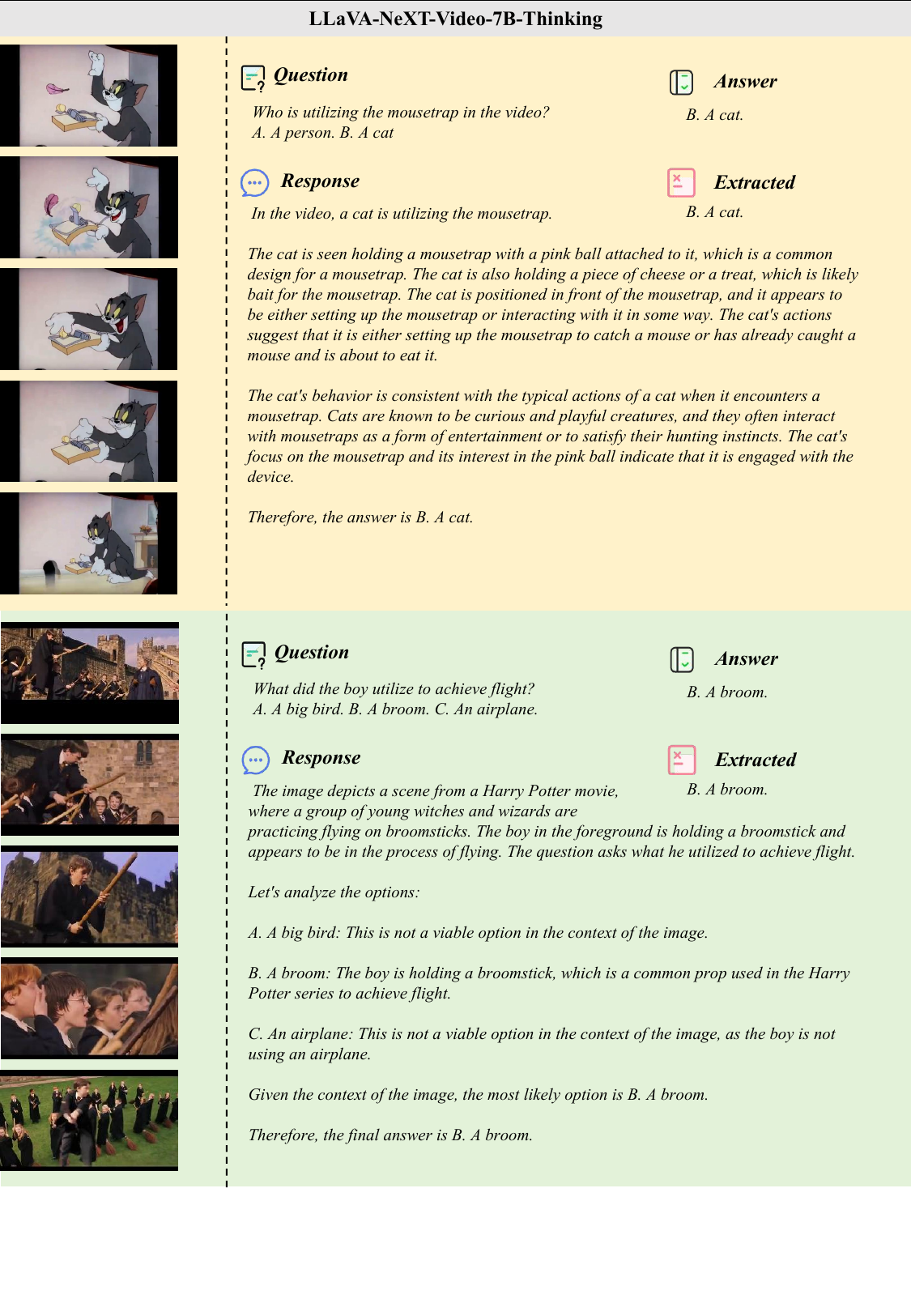}
\end{center}
\vspace{-2.5cm}
   \caption{Case of response from LLaVA-NeXT-Video-7B-Thinking.} 
    \label{ourcase2}
\end{figure*}

Fig.\ref{ourcase2} presents two cases of LLaVA-NeXT-Video-7B-Thinking. By leveraging the proposed thinking-based training strategy, LLaVA-NeXT-Video-7B-Thinking is capable of reasoning and thinking about both video content and problem statements, integrating the semantics of visual and textual modalities to derive the correct results.

\clearpage

\begin{table*}[ht]
\vspace{-0.6cm}
\centering
\begin{tikzpicture}
    \node[draw, fill=gray!10, rounded corners, inner sep=10pt] (box) {
        \begin{minipage}{0.95\textwidth}
            \textbf{\texttt{Short-Answer:}} \\
            
            You are a professional homework grading tool. I will provide you with four rules for grading \\
            1. The Standard Answer is a sentence. Compare the provided predicted answers based on their meaning rather than exact wording. The prediction is correct if it conveys the same intent. \\
            2. If the question asks about identity or species, the predicted answer is correct as long as the core identity or species is correct, even if descriptive adjectives differ. \\
            3. If the question asks about a scene, the predicted answer is correct if the described scene exists in the standard answer or is similar. \\
            4. If the answer indicates that the asked character, object or event is not visible or does not exist, it should be considered as "No answer."  \\
            5. Regardless of the question type, respond only with either 1 or 0, without any additional explanation. \\
            6. 1 means correct, and 0 means incorrect. \\
            For example, if the prediction conveys the same meaning as the standard answer, even if phrased differently, you should respond with 1. \\ 
            Based on the question, is the prediction correct? If yes, return only 1; otherwise, return only 0.  \\
            Question: \texttt{\{question[idx]\}}  \\
            Standard Answer: \texttt{\{answer[idx]\}} \\
            The Predicted Answer: \texttt{\{res[idx]\}} 
        \end{minipage}
    };
\end{tikzpicture}

\caption{The evaluation prompts for Short Answer questions employed in the experiment for models without CoT.}
\label{tab:short_answer_prompt}
\vspace{-0.3cm}
\end{table*}

\begin{table*}[t]

\vspace{-0.3cm}
\centering
\begin{tikzpicture}
    \node[draw, fill=gray!10, rounded corners, inner sep=10pt] (box) {
        \begin{minipage}{0.95\textwidth}
\textbf{\texttt{Multiple-Choice:}} \\

(1) Extracting Answer:  \\
This is a multiple-choice question. Based on the given question and reasoning process, extract the corresponding answer of the reasoning process. \\
                        Question: \texttt{\{question[idx]\}}  \\                  
                        Reasoning Process: \texttt{\{res[idx]\}}  \\                                                   
                        Instructions:  \\
                        1. Identify the correct answer based on the reasoning process. \\
                        2. If the reasoning process directly mentions one of the given choices (A, B, or C), return the corresponding letter along with the full text of that option (e.g., "A. Option text"). \\
                        3. If the reasoning process provides an answer that does not explicitly mention A, B, or C, compare its meaning to the given choices and return the best-matching option in the format "Letter. Option text". \\
                        4. If the reasoning process concludes that the correct answer is "no answer" or "I don't know", return "no answer". \\
                        5. Return only the final answer, without explanation or additional text. \\
                        6. Foucs more on the final summary sentence. \\
                        
(2) Judging: \\
You are a professional homework grading tool. I will provide you with four rules for grading:  \\ 
                    1. This is a multiple-choice question. Judge the correctness based on selected letter and actual content of the provided answers.  \\ 
                    2. Regardless of the question type, respond only with either 1 or 0, without any additional explanation.  \\  
                    3. 1 means the prediction is correct, and 0 means it is incorrect.  \\ 
                    4. If the predicted answer matches the correct answer in meaning, even if it is phrased differently, consider it correct.  \\ 
                    For example, if the prediction conveys the same meaning as the standard answer, you should respond with 1.  \\ 
                    Based on the question and its standard answer, is the prediction correct? If yes, return only 1; otherwise, return only 0.  \\ 
                    Question: \texttt{\{question[idx]\}}  \\
                    Standard Answer: \texttt{\{answer[idx]\}}  \\ 
                    The Predicted Answer: \texttt{\{extract\_answer\}}  
        \end{minipage}
    };
\end{tikzpicture}
\caption{The evaluation prompts for multiple-choice questions employed in the experiment for models with CoT.}
\label{tab:Cot_MC_prompt}
\vspace{-0.65cm}
\end{table*}

\begin{table*}[ht]
\centering
\begin{tikzpicture}
    \node[draw, fill=gray!10, rounded corners, inner sep=10pt] (box) {
        \begin{minipage}{0.95\textwidth}
\textbf{\texttt{Binary-Choice:}} \\

(1) Extracting Answer:  \\
This is a multiple-choice question. Based on the given question and reasoning process, extract the corresponding answer of the reasoning process. \\
                        Question: \texttt{\{question[idx]\}}  \\                  
                        Reasoning Process: \texttt{\{res[idx]\}}  \\                                                   
                        Instructions:  \\
                        1. Identify the correct answer based on the reasoning process. \\
                        2. If the reasoning process explicitly states "yes" or "no", return direct "yes" or "no"  \\
                        3. If the reasoning process concludes that the correct answer is "no answer" or "I don't know", return "no answer". \\
                        5. Return only the final "yes" or "no", without explanation or additional text. \\
                        
(2) Judging: \\
You are a professional homework grading tool. I will provide you with four rules for grading \\
                    1. This is a yes/no question. The Standard Answer is only Yes/No, please directly compare the standard answer with 'yes' or 'no' in the predicted answer. \\
                    2. No matter what kind of questions, only response with one of 1 or 0. No more explaination.  \\
                    3. 1 means correct (they are the same), 0 means wrong (they are different).  \\
                    For example, if the prediction conveys the same meaning as the standard answer, you should respond with 1. \\ 
                    Based on the question, is the prediction correct? If yes, only return 1, otherwise only return 0. \\ 
                    Question: \texttt{\{question[idx]\}}  \\
                    Standard Answer: \texttt{\{answer[idx]\}} \\
                    The Predicted Answer: \texttt{\{extract\_answer\}} 
        \end{minipage}

    };
\end{tikzpicture}
\caption{The evaluation prompts for Binary-Choice questions employed in the experiment for models with CoT.}
\label{tab:Cot_BC_prompt}
\end{table*}

\begin{table*}[ht]
\centering
\begin{tikzpicture}
    \node[draw, fill=gray!10, rounded corners, inner sep=10pt] (box) {
        \begin{minipage}{0.95\textwidth}
\textbf{\texttt{Short-Answer:}} \\

You are a professional homework grading tool. I will provide you with four rules for grading \\
                    1. The Standard Answer is a sentence. Compare the provided predicted answers based on their meaning rather than exact wording. The prediction is correct if it conveys the same intent. \\  
                    2. If the question asks about identity or species, the predicted answer is correct as long as the core identity or species is correct, even if descriptive adjectives differ. \\
                    3. If the question asks about a scene, the predicted answer is correct if the described scene exists in the standard answer or is similar. \\  
                    4. If the answer indicates that the asked character, object or event is not visible or does not exist, it should be considered as "No answer."  \\
                    5. Regardless of the question type, respond only with either 1 or 0, without any additional explanation. \\
                    6. 1 means correct, and 0 means incorrect. \\
                    For example, if the prediction conveys the same meaning as the standard answer, even if phrased differently, you should respond with 1. \\  
                    Based on the question, is the prediction correct? If yes, return only 1; otherwise, return only 0.  \\
                    Question: \texttt{\{question[idx]\}}  \\
                    Standard Answer: \texttt{\{answer[idx]\}} \\
                    The Predicted Answer: \texttt{\{res[idx]\}} \\
        \end{minipage}
    };
\end{tikzpicture}
\caption{The evaluation prompts for Short Answer questions employed in the experiment for models with CoT.}
\label{tab:Cot_SA_prompt}
\end{table*}


\end{document}